\documentclass[twoside,11pt]{article}

\usepackage{color}
\usepackage{jmlr2e}
\usepackage{amsmath}
\usepackage[ansinew]{inputenc}
\usepackage{bm}
\usepackage{float}
\usepackage{url}
\usepackage{algorithm}
\usepackage{algorithmic}
\usepackage{graphics}
\usepackage{subfigure}
\usepackage[small]{caption}

\DeclareMathOperator{\erf}{erf}         
\DeclareMathOperator{\tr}{trace}           
\DeclareMathOperator{\cov}{cov}         
\DeclareMathOperator{\ex}{E}            
\DeclareMathOperator{\bdiag}{blockdiag} 
\DeclareMathOperator{\vecO}{vec}        
\DeclareMathOperator{\diag}{diag}       
\providecommand{\abs}[1]{\lvert#1\rvert}
\newcommand{\boldx}{\mathbf{x}} 
\newcommand{\boldz}{\mathbf{z}} 
\newcommand{\boldK}{\mathbf{K}} 
\newcommand{\boldf}{\mathbf{f}} 
\newcommand{\boldB}{\mathbf{B}} 
\newcommand{\boldA}{\mathbf{A}} 
\newcommand{\boldu}{\mathbf{u}} 
\newcommand{\bolda}{\mathbf{a}} 
\newcommand{\boldAtilde}{\mathbf{\widetilde{A}}} 
\newcommand{\boldUpsi}{\bm{\Upsilon}}
\newcommand{\boldX}{\mathbf{X}} 
\newcommand{\boldI}{\mathbf{I}} 
\newcommand{\inputSpace}{\mathcal{X}} 
\newcommand{\dif}{\textrm{d}}   
\newcommand{\params}{\bm{\theta}} 
\newcommand{\veC}{\textbf{\hspace{-0.001in}:}} 
\newcommand{\preci}{\mathbf{P}}
\newcommand{\dirFigures}{./diagrams/}
\firstpageno{1}

\begin{document}
\title{Sparse Convolved Multiple Output Gaussian Processes}

\author{\name Mauricio A. Álvarez \email alvarezm@cs.man.ac.uk \\
       \name Neil D. Lawrence \email neill@cs.man.ac.uk \\
       \addr School of Computer Science\\
       University of Manchester\\
       Manchester, UK, M13 9PL
}
\maketitle

\begin{abstract}
  Recently there has been an increasing interest in methods that deal
  with multiple outputs. This has been motivated partly by frameworks
  like multitask learning, multisensor networks or structured output
  data. From a Gaussian processes perspective, the problem reduces to
  specifying an appropriate covariance function that, whilst being
  positive semi-definite, captures the dependencies between all the
  data points and across all the outputs. One approach to account for
  non-trivial correlations between outputs employs convolution
  processes. Under a latent function interpretation of the convolution
  transform we establish dependencies between output variables. The
  main drawbacks of this approach are the associated computational and
  storage demands.  In this paper we address these issues. We present
  different sparse approximations for dependent output Gaussian
  processes constructed through the convolution formalism. We exploit
  the conditional independencies present naturally in the model. This
  leads to a form of the covariance similar in spirit to the so called
  PITC and FITC approximations for a single output. We show
  experimental results with synthetic and real data, in particular, we
  show results in pollution prediction, school exams score prediction and gene expression data.
\end{abstract}


\section{Introduction}

Accounting for dependencies between model outputs has important
applications in several areas. In sensor networks, for example,
missing signals from temporal failing sensors may be predicted due to
correlations with signals acquired from other sensors
\citep{Rogers:towards08}. In geostatistics, prediction of the
concentration of heavy pollutant metals (for example, Copper), that
are expensive to measure, can be done using inexpensive and
oversampled variables (for example, pH) as a proxy.  Within the
machine learning community this approach is sometimes known as
multitask learning. The idea in multitask learning is that information
shared between the tasks leads to improved performance in comparison
to learning the same tasks individually \citep{Caruana:MTL:1997}.

In this paper, we consider the problem of modeling related outputs in
a Gaussian process (GP). A Gaussian process specifies a prior
distribution over functions. When using a GP for multiple related
outputs, our purpose is to develop a prior that expresses correlation
between the outputs. This information is encoded in the covariance
function. The class of valid covariance functions is the same as the
class of reproducing kernels.\footnote{In this paper we will use kernel
  to refer to both reproducing kernels and smoothing
  kernels. Reproducing kernels are those used in machine learning that
  conform to Mercer's theorem. Smoothing kernels are kernel functions
  which are convolved with a signal to create a smoothed version of
  that signal.} Such kernel functions for single outputs are widely
studied in machine learning \citep[see, for example,
][]{Rasmussen:book06}. More recently the community has begun to turn
its attention to covariance functions for multiple outputs. One of the
paradigms that has been considered
\citep{Teh:semiparametric05,Rogers:towards08, Bonilla:multi07} is
known in the geostatistics literature as \emph{the linear model of
  coregionalization} (LMC)
\citep{Journel:miningBook78,Goovaerts:book97}.  In the LMC, the
covariance function is expressed as the sum of Kronecker products
between \emph{coregionalization matrices} and a set of underlying
covariance functions. The correlations across the outputs are
expressed in the coregionalization matrices, while the underlying
covariance functions express the correlation between different data
points.

Multitask learning has also been approached from the perspective of
\emph{regularization theory}
\citep{Evgeniou:multitaskSVM04,Evgeniou:multitask05}. These
\emph{multitask kernels} are obtained as generalizations of the
regularization theory to vector-valued functions. They can also be
seen as examples of LMCs applied to linear transformations of the input space. 

The linear model of coregionalization is a rather restrictive approach
to constructing multiple output covariance functions. Each output can
be thought of as an instantaneous mixing of the underlying
signals/processes. An alternative approach to constructing covariance
functions for multiple outputs employs \emph{convolution processes}
(CP). To obtain a CP in the single output case, the output of a given
process is convolved with a smoothing kernel function. For example, a
white noise process may be convolved with a smoothing kernel to obtain
a covariance function \citep{Barry:balckbox96,
  verHoef:convolution98}. \citet{Higdon:convolutions02} noted that if
a single input process was convolved with different smoothing kernels
to produce different outputs, then correlation between the outputs
could be expressed. This idea was introduced to the machine learning
audience by \citet{Boyle:dependent04}. We can think of this approach
to generating multiple output covariance functions as a
non-instantaneous mixing of the base processes. 
    
The convolution process framework is an elegant way for constructing
dependent output processes. However, it comes at the price of having
to consider the full covariance function of the joint GP.  For $D$
output dimensions and $N$ data points the covariance matrix scales as
$DN$ leading to $O(N^3D^3)$ computational complexity and $O(N^2D^2)$
storage.   
We are interested in exploiting the richer class of
covariance structures allowed by the CP framework, but reducing the
additional computational overhead they imply.

In this paper, we propose different sparse approximations for the full
covariance matrix involved in the multiple output convolution process.
We exploit the fact that, in the convolution framework, each of the
outputs is conditional independent of all others if the input process
is fully observed.  This leads to an approximation that turns out to
be strongly related to the partially independent training conditional
(PITC) \citep{Quinonero:unifying05} approximation for a single output
GP. This analogy inspires us to consider a further conditional
independence assumption across data points. This leads to an
approximation which shares the form of the fully independent training
conditional (FITC) approximation
\citep{Snelson:pseudo05,Quinonero:unifying05}. This reduces
computational complexity to $O(NDK^2)$ and storage to $O(NDK)$ with
$K$ representing a user specified value for the number of inducing
points in the approximation.

The rest of the paper is organized as follows. First we give a more
detailed review of related work, with a particular
focus on relating multiple output work in machine learning to other
fields. Despite the fact that there are several other approaches to
multitask learning (see for example \citet{Caruana:MTL:1997,
  Heskes:empiricalBayes:00, Bakker:taskClusteringMTL:2003,
  Xue:MTLDirichlet:2007} and references therein), in this paper, we
focus our attention to those that address the problem of constructing
the covariance or kernel function for multiple outputs, so that it can
be employed, for example, together with Gaussian process prediction.
Then we review the convolution process approach in Section
\ref{section:convolution} and Section \ref{section:examples:kernels}. We demonstrate how our conditional independence
assumptions can be used to reduce the computational load of inference
in Section \ref{section:sparse}. Experimental results
are shown in Section \ref{section:experiments} and finally some
discussion and conclusions are presented in Section
\ref{section:conclusions}.

\section{Related Work}\label{section:related:work}

In geostatistics, multiple output models are used to model the
co-occurrence of minerals or pollutants in a spatial field.  Many of
the ideas for constructing covariance functions for multiple outputs
have first appeared within the geostatistical literature, where they
are known as linear models of coregionalization (LMC). We present the LMC and then 
review how several models proposed in the machine learning 
literature can be seen as special cases of the LMC. 

\subsection{The Linear Model of Coregionalization}

The term linear model of coregionalization refers to models in which
the outputs are expressed as \emph{linear} combinations of independent
random functions. If the independent random functions are Gaussian processes then the resulting model will also be a Gaussian process with a positive semi-definite covariance function.  Consider a
set of $D$ output functions  $\{f_d(\mathbf{x})\}_{d=1}^D$ where $\boldx\in\Re^{p}$ is the input domain. 
In a LMC  each output function, $f_d(\boldx)$, is
expressed as \citep{Journel:miningBook78}
\begin{align}\label{eq:lmc:basic}
f_d(\boldx) &= \sum_{q =1}^Qa_{d,q}u_{q}(\boldx)+\mu_d.
\end{align}
Under the GP interpretation of the LMC the functions $\left\{u_{q}(\boldx)\right\}_{q=1}^Q$ are taken (without loss of generality) to be drawn from a zero-mean GP with  $\cov[u_q(\boldx),u_{q'}(\boldx')]=k_{q}(\boldx,\boldx')$ if $q=q'$ and zero otherwise. Some of these base processes might have the same covariance, i.e. $k_{q}(\boldx,\boldx')=k_{q'}(\boldx,\boldx')$, but they would still be independently sampled. We can group together \citep{Journel:miningBook78,Goovaerts:book97} the base processes that share latent functions, allowing us to express a given output as
\begin{align*}
f_d(\boldx) &= \sum_{q=1}^Q\sum_{i=1}^{R_{q}}a_{d,q}^{i}u_q^{i}(\boldx)+\mu_d,
\end{align*}
where the functions $\left\{u_{q}^{i}(\boldx)\right\}_{i=1}^{R_q}$, $i=1,\ldots, R_q$, 
represent the latent functions that share the same covariance matrix $k_{q}(\boldx,\boldx')$. 
There are now $Q$ groups of functions, each member of a group shares the same covariance, but is sampled independently.

In geostatistics it is common to simplify the analysis of these models
by assuming that the processes $f_d(\boldx)$ are stationary
\citep{Cressie:spatialdataBook:1993}.  The stationarity and ergodicity
conditions are introduced so that the prediction stage can be realized
through an optimal linear predictor using a single realization of the
process \citep{Cressie:spatialdataBook:1993}. Such linear predictors
receive the general name of \emph{cokriging}.  The cross covariance
between any two functions $f_d(\boldx)$ and $f_{d'}(\boldx)$ is given
in terms of the covariance functions for $u_{q}^{i}(\boldx)$
\begin{align}\notag
\cov[f_d(\boldx),f_{d'}(\boldx')]&=\sum_{q=1}^Q\sum_{q'=1}^Q\sum_{i=1}^{R_q}\sum_{i'=1}^{R_q}a_{d, q}^ia_{d',q'}^{i'}
\cov[u_q^{i}(\boldx),u_{q'}^{i'}(\boldx')].
\end{align}
Because of the independence of the latent functions $u_{q}^{i}(\boldx)$, the above expression can be reduced to
\begin{align}\label{eq:lmc:fullCov}
\cov[f_d(\boldx),f_{d'}(\boldx')]&=\sum_{q=1}^Q\sum_{i=1}^{R_q}a_{d,q}^ia_{d',q}^{i}k_{q}(\boldx, \boldx')
=\sum_{q=1}^Q b_{d,d'}^qk_{q}(\boldx, \boldx'),
\end{align}
with $b_{d,d'}^q=\sum_{i=1}^{R_q}a_{d,q}^ia_{d',q}^{i}$. 

For a number $N$ of input vectors, let $\boldf_d$ be the vector of
values from the output $d$ evaluated at
$\boldX=\{\boldx_n\}_{n=1}^{N}$. If each output has the same set of
inputs the system is known as \emph{isotopic}.  In general, we can
allow each output to be associated with a different set of inputs,
$\boldX^{(d)}=\{\boldx^{(d)}_n\}_{n=1}^{N_d}$, this is known as
\emph{heterotopic}.\footnote{These names come from geostatistics.} For notational simplicity, we restrict ourselves to
the isotopic case, but our analysis can also be completed for
heterotopic set ups. The covariance matrix for $\boldf_d$ is obtained
expressing equation \eqref{eq:lmc:fullCov} as
\begin{align*}
\cov[\boldf_d,\boldf_{d'}]&= \sum_{q=1}^Q\sum_{i=1}^{R_q}a_{d,q}^ia_{d',q}^{i}\boldK_q=\sum_{q=1}^Qb^q_{d,d'}\boldK_q ,
\end{align*}
where $\boldK_q\in\Re^{N\times N}$ has entries given by computing  $k_q(\boldx, \boldx')$ for all combinations from $\boldX$. We now define $\boldf$ to be a stacked version of the
outputs so that $\boldf =
[\boldf_1^{\top},\ldots,\boldf_D^{\top}]^{\top}$. We can now write the covariance matrix for the joint process over $\boldf$  as
\begin{align}\label{eq:matricesLMC}
\boldK_{\boldf,\boldf} &= \sum_{q=1}^Q\boldA_{q}\boldA^{\top}_{q} \otimes \boldK_q = \sum_{q=1}^Q\boldB_q \otimes \boldK_q,
\end{align}
where the symbol $\otimes$ denotes the Kronecker product,
$\boldA_q\in\Re^{D\times R_q}$ has entries $a_{d,q}^i$ and
$\boldB_q=\boldA_{q}\boldA^{\top}_{q} \in\Re^{D\times D}$ has entries
$b_{d,d'}^{q}$ and is known as the \emph{coregionalization matrix}.
The covariance matrix $\boldK_{\boldf,\boldf}$ is positive
semi-definite as long as the coregionalization matrices $\boldB_q$ are
positive semi-definite and $k_q(\boldx, \boldx')$ is a valid covariance
function. By definition, coregionalization matrices
$\boldB_q$ fulfill the positive semi-definiteness requirement. The
covariance functions for the latent processes, $k_q(\boldx, \boldx')$,
can simply be chosen from the wide variety of covariance functions
(reproducing kernels) that are used for the single output
case. Examples include the squared exponential (sometimes called the
Gaussian kernel or RBF kernel) and the Mat\'{e}rn class of covariance
functions \citep[see][chap.~4]{Rasmussen:book06}.

The linear model of coregionalization represents the covariance
function as a product of the contributions of two covariance
functions. One of the covariance functions models the dependence
between the functions independently of the input vector $\boldx$, this
is given by the coregionalization matrix $\boldB_q$, whilst the other
covariance function models the input dependence independently of the
particular set of functions $f_d(\boldx)$, this is the covariance
function $k_q(\boldx, \boldx')$.

We can understand the LMC by thinking of the functions having been
generated as a two step process. Firstly we sample a set of
independent processes from the covariance functions given by
$k_q(\boldx, \boldx')$, taking $R_q$ independent samples for each
$k_q(\boldx,\boldx')$. We now have $R=\sum_{q=1}^QR_q$ independently sampled
functions. These functions are \emph{instantaneously
  mixed}\footnote{The term instantaneous mixing is taken from blind
  source separation. Of course if the underlying processes are not
  temporal but spatial, instantaneous is not being used in its
  original sense. However, it allows us to distinguish this mixing
  from convolutional mixing.} in a linear fashion. In other words the
output functions are derived by application of a scaling and a
rotation to an output space of dimension $D$.

\subsubsection{Intrinsic Coregionalization Model }

A simplified version of the LMC, known as the intrinsic
coregionalization model (ICM) \citep[see][]{Goovaerts:book97}, assumes
that the elements $b_{d,d'}^q$ of the coregionalization matrix
$\boldB_q$ can be written as $b_{d,d'}^q=\upsilon_{d,d'}b_q$.  In other
words, as a scaled version of the elements $b_q$ which do not depend
on the particular output functions $f_d(\boldx)$.  Using this form for
$b_{d,d'}^q$, equation \eqref{eq:lmc:fullCov} can be expressed as
\begin{align*}
\cov[f_d(\boldx),f_{d'}(\boldx')]&=\sum_{q=1}^Q\upsilon_{d,d'}b_qk_{q}(\boldx, \boldx')
=\upsilon_{d,d'}\sum_{q=1}^Q b_qk_{q}(\boldx,\boldx').
\end{align*}
The covariance matrix for $\boldf$ takes the form
\begin{align}\label{eq:multiOut:icm:fullCov:matrixForm}
\boldK_{\boldf,\boldf}&= \boldUpsi \otimes \boldK,
\end{align}
where $\boldUpsi\in\Re^{D\times D}$ with entries $\upsilon_{d,d'}$ and
$\boldK=\sum_{q=1}^Qb_q\boldK_q$ is an equivalent valid covariance
function. This is also equivalent to a LMC model where we have
$Q=1$. As pointed out by \citet{Goovaerts:book97}, the ICM is
much more restrictive than the LMC since it assumes that each basic
covariance $k_q(\boldx, \boldx')$ contributes equally to the
construction of the autocovariances and cross covariances for the
outputs.

\subsubsection{Linear Model of Coregionalization in Machine Learning}

Several of the approaches to multiple output learning in machine learning based on kernels can be seen 
as examples of the linear model of coregionalization. 

\paragraph{Semiparametric latent factor model.} In \citet{Teh:semiparametric05}, the model proposed to construct the covariance
function for multiple outputs turns out to be a simplified version of equation 
\eqref{eq:matricesLMC}. In particular, if $R_q=1$ (see equation \eqref{eq:lmc:basic}), we can rewrite 
equation \eqref{eq:matricesLMC} as
\begin{align}\notag
\boldK_{\boldf,\boldf} &= \sum_{q=1}^Q\bolda_{q}\bolda^{\top}_{q} \otimes \boldK_q,
\end{align}
where $\bolda_{q}\in\Re^{D\times 1}$ with elements $a_{d,q}$. With some algebraic manipulations that exploit the properties of the Kronecker product\footnote{See \citet{Brookes:matrix05} for a nice overview.} we can write
\begin{align}\notag
\boldK_{\boldf,\boldf} &= \sum_{q=1}^Q(\bolda_{q}\otimes \boldI_N)\boldK_q(\bolda^{\top}_{q} \otimes \boldI_N)
=(\boldAtilde \otimes \boldI_N)\mathbf{\widetilde{K}}(\boldAtilde^{\top} \otimes \boldI_N),
\end{align}
where $\boldI_N$ is the $N$-dimensional identity matrix, $\boldAtilde\in\Re^{D\times Q}$ is a matrix 
with columns $\bolda_q$ and $\mathbf{\widetilde{K}}\in\Re^{QN\times QN}$ 
is a block diagonal matrix with blocks given by $\boldK_q$. 

The functions $u_q(\boldx)$ are considered to be latent factors and
the model for the outputs was named semiparametric latent factor model
(SLFM). The semiparametric name comes from the fact that it is
combining a nonparametric model, \emph{i.e.} a Gaussian process with a
parametric linear mixing of the functions $u_{q}(\boldx)$. The kernel
for each basic process $q$, $k_q(\boldx,\boldx')$, is assumed to be of Gaussian type with
a different length scale per input dimension.  For computational speed up the informative
vector machine (IVM) is employed \citep{Lawrence:ivm02}.

\paragraph{Multi-task Gaussian processes.} The intrinsic
coregionalization model has been employed in \citet{Bonilla:multi07}
for multitask learning. The covariance matrix is expressed as
$\boldK_{\overline{\boldf}(\boldx),\overline{\boldf}(\boldx')}=
K^f\otimes k(\mathbf{x},\mathbf{x'})$, with $\overline{\boldf}(\boldx)=[f_1(\boldx),\ldots,
f_D(\boldx)]^{\top}$, $K^f$
being constrained positive semi-definite and
$k(\mathbf{x},\mathbf{x'})$ a covariance function over inputs. It can
be noticed that this expression has is equal to
the one in \eqref{eq:multiOut:icm:fullCov:matrixForm}, when it is evaluated for $\boldx,\boldx'\in\boldX$. In
\citet{Bonilla:multi07}, $K^f$ (equal to $\boldUpsi$ in equation 
\eqref{eq:multiOut:icm:fullCov:matrixForm}) expresses the correlation between
tasks or inter-task dependencies and it is represented through a
\textrm{PPCA} model, with the spectral factorization in the
\textrm{PPCA} model replaced by an incomplete Cholesky
decomposition to keep numerical stability. For $k(\mathbf{x},\mathbf{x'})$ (the function used in equation 
\eqref{eq:multiOut:icm:fullCov:matrixForm} to compute $\boldK$), the
squared-exponential kernel is employed. To reduce computational
complexity, the Nystr\"{o}m approximation is applied.

It can be shown that if the outputs are considered to be noise-free,
prediction using the intrinsic coregionalization model under an
isotopic data case is equivalent to independent prediction over each
output \citep{Helterbrand:universalCR94}.  This circumstance is also
known as autokrigeability \citep{Wackernagel:book03} and it can also
be seen as the cancellation of inter-task transfer
\citep{Bonilla:multi07}.

\paragraph{Multi-output Gaussian processes.} The intrinsic
coregionalization model has been also used in
\citet{Rogers:towards08}. Matrix $\boldUpsi$ in
expression \eqref{eq:multiOut:icm:fullCov:matrixForm} is assumed to be
of the spherical parametrisation kind, $\boldUpsi =
\diag(\mathbf{e})\mathbf{S} ^{\top}\mathbf{S}\diag(\mathbf{e})$, where
$\mathbf{e}$ gives a description for the length scale of each output
variable and $\mathbf{S}$ is an upper triangular matrix whose $i$-th
column is associated with particular spherical coordinates of points
in $\Re^i$ \citep[for details
see][sec.~3.4]{osborne:multiGPsReport:2007}. Function $k(\boldx,
\boldx')$ is represented through a Mátern kernel, where different
parametrisations of the covariance allow the expression of periodic
and non-periodic terms.  Sparsification for this model is obtained
using an IVM style approach.

\paragraph{Multi-task kernels.} Kernels for multiple outputs have also been studied in the context of
regularization theory. The approach is based mainly on the definition
of kernels for multitask learning provided in
\citet{Evgeniou:multitaskSVM04} and \citet{Evgeniou:multitask05},
derived based on the theory of kernels for vector-valued
functions. Let $\mathcal{D}=\{1,\ldots,D\}$. According to
\citet{Evgeniou:multitask05}, the following lemma can be used to
construct multitask kernels,
\begin{lemma}\label{lemma:MTL}
If $G$ is a kernel on $\mathcal{T}\times\mathcal{T}$ and, for every $d\in \mathcal{D}$
there are prescribed mappings
$\Phi_d:\inputSpace\rightarrow\mathcal{T}$ such that
\[
k_{d,d'}(\boldx, \boldx')=k((\mathbf{x},d),(\mathbf{x}',d')) = G(\Phi_d(\mathbf{x}),\Phi_{d'}(\mathbf{x}')),\quad \mathbf{x},
  \mathbf{x}'\in\Re^p,\;d,d'\in\mathcal{D},
\]
then $k(\cdot)$ is a multitask or multioutput kernel.
\end{lemma}
A linear multitask kernel can be obtained if we set $\mathcal{T}=\Re^m$, $\Phi_d(\boldx)=\mathbf{C}_d\boldx$ with
$\Phi_d\in\Re^{m\times p}$ and $G:\Re^m\times\Re^m\rightarrow\Re$ as the polynomial kernel $G(\mathbf{z},\mathbf{z}')=
(\mathbf{z}^{\top}\mathbf{z}')^n$ with $n=1$, leading to   
$k_{d,d'}(\boldx, \boldx') = \boldx^{\top}\mathbf{C}^{\top}_d\mathbf{C}_{d'}\boldx'$.
Lemma \ref{lemma:MTL} can be seen as the result of applying kernel
properties to the mapping $\Phi_d(\boldx)$
\citep[see][pag.~2]{Genton:kernelsStatistics:2001}. Notice that this corresponds to the linear model of 
coregionalization where each output is expressed through its own basic process acting over the linear transformation 
$\mathbf{C}_d\boldx$, this is, $u_d(\Phi_d(\boldx))=u_d(\mathbf{C}_d\boldx)$.  

\section{Convolution processes for multiple outputs}\label{section:convolution}

The approaches introduced above all involve some form of instantaneous
mixing of a series of independent processes to construct correlated
processes. Instantaneous mixing has some limitations. If we wanted to
model two output processes in such a way that one process was a
blurred version of the other, we cannot achieve this through
instantaneous mixing. We can achieve blurring through convolving a
base process with a smoothing kernel. If the base process is a
Gaussian process, it turns out that the convolved process is also a
Gaussian process. We can therefore exploit convolutions to construct
covariance functions \citep{Barry:balckbox96,verHoef:convolution98,
  Higdon:ocean98,Higdon:convolutions02}. A recent review of several
extensions of this approach for the single output case is presented in
\citet{Calder:convolution07}. Applications include the construction of
nonstationary covariances
\citep{Higdon:ocean98,Higdon:nonstationaryCov:1998,Fuentes:nonstationaryAirPollution:2002,
  Fuentes:spectralNonstationary:2002,Pacioreck:nonstationaryCov:2004}
and spatiotemporal covariances
\citep{Wikle:hierarchicalBayesSpaceTimeModels98,Wikle:kernelBasedSpectralModel02,Wikle:hierarchicalEcologicalProcesses03}.

\citet{Higdon:convolutions02} suggested using convolutions to construct multiple output covariance functions. The approach was introduced to the machine learning community by \citet{Boyle:dependent04}. Consider again a set of $D$ functions $\{f_d(\mathbf{x})\}_{d=1}^D$. Now each function could be expressed through a convolution integral between a smoothing kernel, 
$\{G_{d}(\mathbf{x})\}_{d=1}^D$, and a latent function
$u(\mathbf{x})$,
\begin{equation}
f_d(\mathbf{x}) = \int_{\inputSpace}
G_{d}(\mathbf{x}-\mathbf{z})u(\mathbf{z})\dif\mathbf{z}.\label{eq:convolution}
\end{equation}
More generally, we can consider the influence of more than one latent
function, $\{u_q(\mathbf{z})\}_{q=1}^Q$, and corrupt each of the outputs of the convolutions 
with an independent process (which could also include a noise term), $w_d(\mathbf{x})$, to obtain
\begin{align}
y_d(\mathbf{x}) =  f_d(\mathbf{x}) + w_d(\mathbf{x})
=\sum_{q=1}^Q\int_{\inputSpace}
G_{d,q}(\mathbf{x}-\mathbf{z})u_q(\mathbf{z})\dif\mathbf{z} +
w_d(\mathbf{x}).\label{eq:process:conv}
\end{align}
The covariance between two different outputs $y_d(\mathbf{x})$ and $y_{d'}(\mathbf{x'})$ is then recovered as
\begin{align}
\cov\left[y_d(\mathbf{x}),y_{d'}(\mathbf{x'})\right] = &
\cov\left[f_d(\mathbf{x}),f_{d'}(\mathbf{x'})\right] +
\cov\left[w_d(\mathbf{x}),w_{d'}(\mathbf{x'})\right]\delta_{d,d'},\nonumber
\end{align}
where $\delta_{d,d'}$ is the Kronecker delta function and
\begin{align}
\cov\left[f_d(\mathbf{x}),f_d'(\mathbf{x'})\right]= &
\sum_{q=1}^Q\sum_{q'=1}^Q\int_{\inputSpace}
G_{d,q}(\mathbf{x}-\mathbf{z})\int_{\inputSpace}
G_{d',q'}(\mathbf{x'}-\mathbf{z'})\cov\left[u_q(\mathbf{z}),u_{q'}(\mathbf{z'})\right]\dif\mathbf{z'}\dif\mathbf{z}.
\label{eq:covf}
\end{align}
Specifying $G_{d,q}(\boldx-\mathbf{z})$ and
$\cov\left[u_q(\mathbf{z}),u_{q'}(\mathbf{z'})\right]$ in
\eqref{eq:covf}, the covariance for the outputs $f_d(\mathbf{x})$ can
be constructed indirectly. Note that if the smoothing kernels are taken to be the Dirac delta function such that,
\[
G_{d,q}(\mathbf{x}-\mathbf{z}) = a_{d,q}\delta(\mathbf{x}-\mathbf{z'}),
\]
where $\delta(\cdot)$ is the Dirac delta function,\footnote{We have
  slightly abused of the delta notation to indicate the Kronecker
  delta for discrete arguments and the Dirac function for continuous
  arguments. The particular meaning should be understood from the
  context.} the double integral is easily solved and the linear model
of coregionalization is recovered. This matches to the concept of
\emph{instantaneous mixing} we introduced to describe the LMC. In a
convolutional process the mixing is more general, for example the
latent process could be smoothed for one output, but not smoothed for
another allowing correlated output functions of different length
scales.

The traditional approach to convolution processes in statistics and
signal processing is to assume that the latent functions
$u_q(\mathbf{z})$ are independent white Gaussian noise processes,
$\cov\left[u_q(\mathbf{z}),u_{q'}(\mathbf{z'})\right]=\sigma^2_{u_q}\delta_{q,q'}\delta(\mathbf{z}-\mathbf{z'})$.
This allows us to simplify \eqref{eq:covf} as
\begin{equation*}
\cov\left[f_d(\mathbf{x}),f_{d'}(\mathbf{x'})\right]  =
\sum_{q=1}^Q\sigma^2_{u_q} \int_{\inputSpace}
G_{d,q}(\mathbf{x}-\mathbf{z})G_{d',q}(\mathbf{x'}-\mathbf{z})\dif\mathbf{z}.
\end{equation*}
In general though, we can consider any type of latent process, for example, we
could assume independent GPs for the latent functions so that we have
$\cov\left[u_q(\mathbf{z}),u_{q'}(\mathbf{z'})\right]=k_{u_q,u_{q'}}(\mathbf{z},\mathbf{z'})\delta_{q,q'}$.
With this form of the latent functions, $\eqref{eq:covf}$ can be
written as
\begin{equation}
\cov\left[ f_d(\mathbf{x}),f_{d'}(\mathbf{x'})\right] = \sum_{q=1}^Q
\int_{\inputSpace}
G_{d,q}(\mathbf{x}-\mathbf{z})\int_{\inputSpace}
G_{d',q}(\mathbf{x'}-\mathbf{z'})k_{u_q,u_q}(\mathbf{z},\mathbf{z'})\dif\mathbf{z'}\dif\mathbf{z}.\label{eq:covy:gp}
\end{equation}
As well as this correlation across outputs, the correlation between
the latent function, $u_q(\mathbf{z})$, and any given output,
$f_d(\mathbf{x})$, can be computed,
\begin{equation}
\cov\left[ f_d(\mathbf{x}),u_q(\mathbf{z}))\right] =
\int_{\inputSpace}
G_{d,q}(\mathbf{x}-\mathbf{z'})k_{u_q,u_q}(\mathbf{z'},\mathbf{z})\dif\mathbf{z'}.\label{eq:covy:yu}
\end{equation}

\citet{Higdon:convolutions02} proposed the direct use of convolution
processes for constructing multiple output Gaussian
processes. \citet{Lawrence:gpsim2007a} arrive at a similar
construction from solving a physical model: a first order differential
equation \citep[see also][]{Gao:latent08}. This idea of using physical
models to inspire multiple output systems has been further extended in
\cite{Alvarez:lfm09} who give examples using the heat equation and a
second order system. A different approach using Kalman Filtering ideas
has been proposed in
\citet{Calder:thesis03,Calder:kalmanConvolution07}.  Calder proposed a
model that incorporates dynamical systems ideas to the process
convolution formalism. Essentially, the latent processes are of two
types: random walks and independent cyclic second-order
autoregressions. With this formulation, it is possible to construct a
multivariate output process using convolutions over these latent
processes. Particular relationships between outputs and latent
processes are specified using a special transformation matrix ensuring
that the outputs are invariant under invertible linear transformations
of the underlying factor processes (this matrix is similar in spirit
to the sentitivity matrix of \citet{Lawrence:gpsim2007a} and it is
given a particular form so that not all latent processes affect the
whole set of outputs \citep{Calder:kalmanConvolution07}).

\paragraph{Bayesian kernel methods.} The convolution process is
closely related to the Bayesian kernel method
\citep{Pillai:kernelHilbert07,Liang:bayesianKernelMethods09} for
constructing reproducible kernel Hilbert spaces (RKHS), assigning
priors to signed measures and mapping these measures through integral
operators. In particular, define the following space of functions,
\begin{align*}
\mathcal{F}&=\Big\{f\Big|f(x)=\int_{\inputSpace}G(x,z)\gamma(\dif z),\;\gamma\in\Gamma\Big\},
\end{align*}
for some space $\Gamma\subseteq\mathcal{B}(\inputSpace)$ of signed
Borel measures. In \citet[proposition~1] {Pillai:kernelHilbert07}, the
authors show that for $\Gamma=\mathcal{B}(\inputSpace)$, the space of
all signed Borel measures, $\mathcal{F}$ corresponds to a
RKHS. Examples of these measures that appear in the form of stochastic
processes include Gaussian processes, Dirichlet processes and Lévy
processes.  This framework can be extended for the multiple output
case, expressing the outputs as
\[
f_d(x)=\int_{\inputSpace}G_d(x,z)\gamma(\dif z).
\]
The analysis of the mathematical properties of such spaces of
functions are beyond the scope  of this paper and are postponed for
future work.

\section{Constructing Multiple Output Convolution Processes}\label{section:examples:kernels}

We now consider practical examples of how these multiple output convolution processes can be constructed. We start with a more generic example (although it has an underlying physical interpretation \citep{Alvarez:lfm09}), which can be seen as the equivalent of the squared exponential covariance function for multiple outputs. We will then consider a particular physical model: a simple first order differential equation for modeling transcription.

\paragraph{Example 1. A general purpose convolution kernel for
  multiple outputs.} A simple general purpose kernel for multiple
outputs based on the convolution integral can be constructed assuming
that the kernel smoothing function, $G_{d,r}(\boldx)$, and the
covariance for the latent function, $k_{u_q,u_q}(\boldx, \boldx')$,
follow both a Gaussian form. The kernel smoothing function is given as
\begin{align*}
G_{d,q}(\boldx)&=S_{d,q}\mathcal{N}(\boldx|\mathbf{0},\preci_{d}^{-1}),
\end{align*}
where $S_{d,q}$ is a variance coefficient that depends both of the
output $d$ and the latent function $q$ and $\preci_{d}$ is the
precision matrix associated to the particular output $d$. The
covariance function for the latent process is expressed as
\begin{align*}
k_{u_q,u_q}(\boldx, \boldx')&=\mathcal{N}(\boldx-\boldx'|\mathbf{0}, \bm{\Lambda}_{q}^{-1}),
\end{align*}
with $\bm{\Lambda}_{q}$ the precision matrix of the latent function $q$.

Expressions for the kernels are obtained applying systematically the identity
for the product of two Gaussian distributions. Let $\mathcal{N}(\boldx|\bm{\mu},\preci^{-1})$ denotes a Gaussian for 
$\boldx$, then  
\begin{align}\label{eq:prod:Gaussians}
\mathcal{N}(\boldx|\bm{\mu}_1,\preci_1^{-1})\mathcal{N}(\boldx|\bm{\mu}_2,\preci_2^{-1}) =
\mathcal{N}(\bm{\mu}_1|\bm{\mu}_2,\preci_1^{-1}+\preci_2^{-1})\mathcal{N}(\boldx|\bm{\mu}_c,\preci^{-1}_c), 
\end{align}
where $\bm{\mu}_c=\left(\preci_1+\preci_2\right)^{-1}\left(\preci_1\bm{\mu}_1
+\preci_2\bm{\mu}_2\right)$ and $\preci^{-1}_c=\left(\preci_1+\preci_2\right)^{-1}$. For all integrals we assume 
that $\inputSpace=\Re^p$. 
Using these forms for $G_{d,q}(\boldx)$ and $k_{u_q, u_q}(\boldx, \boldx')$, expression  
\eqref{eq:covy:gp} can be written as  
\begin{align*}
k_{f_d,f_{d'}}(\mathbf{x},\mathbf{x'})& = \sum_{q=1}^QS_{d,q}S_{d',q}
\int_{\inputSpace}
\mathcal{N}(\boldx-\mathbf{z}|\mathbf{0},\preci_{d}^{-1})\int_{\inputSpace}
\mathcal{N}(\boldx'-\mathbf{z'}|\mathbf{0},\preci_{d'}^{-1})
\mathcal{N}(\mathbf{z}-\mathbf{z}'|\mathbf{0},\bm{\Lambda}_q^{-1})\dif\mathbf{z'}\dif\mathbf{z}.
\end{align*}
Since the Gaussian covariance is stationary and isotropic, we can write it as
$\mathcal{N}(\boldx-\boldx'|\mathbf{0},\preci^{-1})=\mathcal{N}(\boldx'-\boldx|\mathbf{0},\preci^{-1})=
\mathcal{N}(\boldx|\boldx',\preci^{-1})=\mathcal{N}(\boldx'|\boldx,\preci^{-1})$. Using the identity in equation 
\eqref{eq:prod:Gaussians} twice, we get
\begin{align*}
k_{f_d,f_{d'}}(\mathbf{x},\mathbf{x'})& = \sum_{q=1}^QS_{d,q}S_{d',q}
\mathcal{N}(\boldx-\boldx'|\mathbf{0},\preci_{d}^{-1}+\preci_{d'}^{-1}+\bm{\Lambda}_{q}^{-1}).
\end{align*}
For a high value of the input dimension, $p$, the term
$1/[(2\pi)^{p/2}|\preci_{d}^{-1}+\preci_{d'}^{-1}+
\bm{\Lambda}_{q}^{-1}|^{1/2}]$ in each of the Gaussian's normalization
terms will dominate, making values go quickly 
to zero.  We can fix this problem, by scaling the outputs using  the factors
$1/[(2\pi)^{p/4}|2\preci_{d}^{-1}+ \bm{\Lambda}_{q}^{-1}|^{1/4}]$ and
$1/[(2\pi)^{p/4}|2\preci_{d'}^{-1}+\bm{\Lambda}_{q}^{-1}|^{1/4}]$. 
Each of these scaling factors correspond to the standard deviation associated to $k_{f_d,f_{d}}(\boldx,\boldx)$ 
and $k_{f_{d'},f_{d'}}(\boldx,\boldx)$.

Equally for the covariance $\cov\left[ f_d(\mathbf{x}),u_q(\mathbf{x}'))\right]$ in equation \eqref{eq:covy:yu}, 
we obtain
\begin{align*}
k_{f_d,u_q}(\mathbf{x},\mathbf{x'})& = S_{d,q}\mathcal{N}(\boldx-\boldx'|\mathbf{0},\preci_{d}^{-1}+\bm{\Lambda}_{q}^{-1}).
\end{align*}
Again, this covariance must be standardized when working in higher dimensions.

\paragraph{Example 2. Convolution kernels constructed through a first order differential equation.} 
The convolution integral appears naturally when solving
ordinary differential equations. In this case, the smoothing kernel function corresponds to the impulse response of the 
system described by the differential equation. As an example consider the following set of first order differential 
equations where the input variable is time, $t$,
\begin{align*}
\frac{\dif f_d(t)}{\dif t}&=\sum_{q=1}^QS_{d,q}u_q(t)-\gamma_df_d(t), \quad d=1,\ldots, D, 
\end{align*}
where $\gamma_d$ is a parameter of the particular system (electrical circuit, mechanical system, among others) and $S_{d,q}$
quantifies the influence of latent function $q$ over output $d$. Assuming initial conditions equal to zero, 
the solution to the above equation is given as 
\begin{align}\label{eq:conv:ode1}
f_d(t)&=\sum_{q=1}^QS_{d,q}\int_{0}^t\exp(-\gamma_d(t-\tau))u_q(\tau)\dif \tau.
\end{align}
If the functions $\{u_q(t)\}_{q=1}^{Q}$ are Gaussian processes with squared exponential kernel 
\begin{align*}
k_{u_q,u_q}(t,t')&= \exp\left(-\frac{(t-t')^2}{\ell^2_q}\right), 
\end{align*}
where $\ell_q$ represents the length-scale parameter, the covariance for the outputs can be found \citep{Lawrence:gpsim2007a,Gao:latent08} as  
\begin{align*}
  k_{f_d,f_{d'}}(t,t')&=
  \sum_{q=1}^QS_{d,q}S_{d',q}\exp(-\gamma_dt-\gamma_{d'}t')\int_{0}^t\exp(\gamma_d\tau)\int_{0}^{t'}
  \exp(\gamma_{d'}\tau')k_{u_q,u_q}(\tau, \tau')\dif \tau \dif\tau'.
\end{align*}
Solving the above equation, the covariance $k_{f_d, f_{d'}}(t,t')$ is given as
\begin{align*}
  k_{f_d,f_{d'}}(t,t')&=\sum_{q=1}^Q\frac{S_{d,q}S_{d',q}\sqrt{\pi}\ell_q}{2}\exp(-\gamma_dt-\gamma_{d'}t')
[h_q(\gamma_{d'},\gamma_{d},t,t')+h_q(\gamma_{d},\gamma_{d'},t',t)],
\end{align*}    
where 
\begin{align*}
h_q(\gamma_{d},\gamma_{d'},t,t')&=\frac{\exp\Big[\Big(\frac{\gamma_d\ell_q}{2}\Big)^2\Big]}{\gamma_d+\gamma_{d'}}\bigg[
\exp[(\gamma_d+\gamma_{d'})t]\mathcal{H}_q(\gamma_d,t,t') - \mathcal{H}_q(\gamma_d,0,t')\bigg],
\end{align*}
and 
\begin{align*}
\mathcal{H}_q(\gamma_d,t,t')&=\erf\bigg(\frac{t}{\ell_q}+\frac{\gamma_d\ell_q}{2}\bigg)-\erf\bigg(\frac{t-t'}{\ell_q}
+\frac{\gamma_d\ell_q}{2}\bigg).
\end{align*}
In the above expression, the function $\erf(x)$ is the so called \emph{error function} and it is defined as 
$\erf(z)=\frac{2}{\sqrt{\pi}}\int_{0}^z\exp(-\xi^2)\dif \xi$. 
The covariance between $f_d(t)$ and $u_q(t')$ is given as
\begin{align*}
k_{f_d,u_q}(t,t')&=\frac{S_{d,q}\sqrt{\pi}\ell_q}{2}\exp\Big[\Big(\frac{\gamma_d\ell_q}{2}\Big)^2\Big]\exp[-\gamma_d(t-t')]
\mathcal{H}_q(\gamma_d,t',t).
\end{align*}

\section{Sparse Approximations for Convolutional Processes}\label{section:sparse}

Assuming that the double integral is tractable, the principle
challenge for the convolutional framework is computing the inverse of
the covariance matrix associated with the outputs. For $D$ outputs,
each having $N$ data points, the inverse has computational complexity
$O(D^3N^3)$ and associated storage of $O(D^2N^2)$. 
We show how
through making specific conditional independence assumptions, inspired
by the model structure \citep{Alvarez:sparse2009}, we arrive at a sparse approximation similar in
form to the partially independent training conditional model
\citep[PITC, see][]{Quinonero:unifying05}. The relationship with PITC
inspires us to make further conditional independence assumptions.

\subsection{Full Dependence}

Given the convolution formalism, we can construct a full GP over the
set of outputs. The likelihood of the model is given by
\begin{equation}
p(\mathbf{y}|\mathbf{X},\bm{\theta}) = \mathcal{N}(\mathbf{0},
\mathbf{K_{f,f}}+\bm{\Sigma}),\label{eq:marginal:full}
\end{equation}
where $\mathbf{y}=\left[\mathbf{y}_1^\top,\ldots,\mathbf{y}_D^\top
\right]^\top$ is the set of output functions with
$\mathbf{y}_d=\left[y_d(\mathbf{x}_1),\ldots,y_d(\mathbf{x}_{N})\right]^\top$;
$\mathbf{K_{f,f}}\in \Re^{DN\times DN}$ is the covariance matrix
arising from the convolution. It expresses the covariance of each data point at every other output and 
data point and its elements are given by
$\cov\left[f_d(\mathbf{x}),f_{d'}(\mathbf{x'})\right]$ in
\eqref{eq:covy:gp}. The term $\boldsymbol{\Sigma}$ represents the covariance associated with the 
independent processes in (\ref{eq:process:conv}), $w_d(\mathbf{x})$. It could contain structure, 
or alternatively could simply represent noise that is independent across the data points. The vector $\bm{\theta}$ refers
to the hyperparameters of the model. 
For exposition we will focus on the isotopic case (although our implementations allow heterotopic modeling), 
so we have a matrix $\mathbf{X}=\{\mathbf{x}_1,\ldots,\mathbf{x}_N\}$ which is the common set of training input
vectors at which the covariance is evaluated.

The predictive distribution for a new set of input vectors
$\mathbf{X}_{*}$ is \citep{Rasmussen:book06}
\[
p(\mathbf{y}_{*}|\mathbf{y},\mathbf{X},\mathbf{X}_*,\bm{\theta}) =
\mathcal{N}\left(\mathbf{K_{f_*,f}}(\mathbf{K_{f,f}}+\bm{\Sigma})^{-1}\mathbf{y},
\mathbf{K_{f_*,f_*}}-\mathbf{K_{f_*,f}}(\mathbf{K_{f,f}}+\bm{\Sigma})^{-1}\mathbf{K_{f,f_*}}+\bm{\Sigma}\right),
\]
where we have used $\mathbf{K_{f_*,f_*}}$ as a compact notation to
indicate when the covariance matrix is evaluated at the inputs
$\mathbf{X}_{*}$, with a similar notation for
$\mathbf{K_{f_*,f}}$. Learning from the log-likelihood involves the
computation of the inverse of $ \mathbf{K_{f,f}}+\bm{\Sigma}$ giving
the problematic complexity of $O(N^3D^3)$. Once the parameters have
been learned, prediction is $O(ND)$ for the predictive mean and
$O(N^2D^2)$ for the predictive variance.

\subsection{Latent functions as conditional means}

We restrict the analysis of the approximations to one latent function $u(\boldx)$. The key to all approximations is based 
on the form we assume for the latent functions. From the perspective of a generative model, equation 
\eqref{eq:convolution} can be interpreted as follows: first we draw
a sample from the Gaussian process prior $p(u(\boldz))$ and then solve the integral for each of the outputs $f_d(\boldx)$
involved. Uncertainty about $u(\boldz)$ is also propagated through the convolution transform. 

\begin{figure*}[ht!]
  \begin{center}
  \subfigure[Conditional prior for $K=5$ ]{ 
  \includegraphics[width=0.30\textwidth]{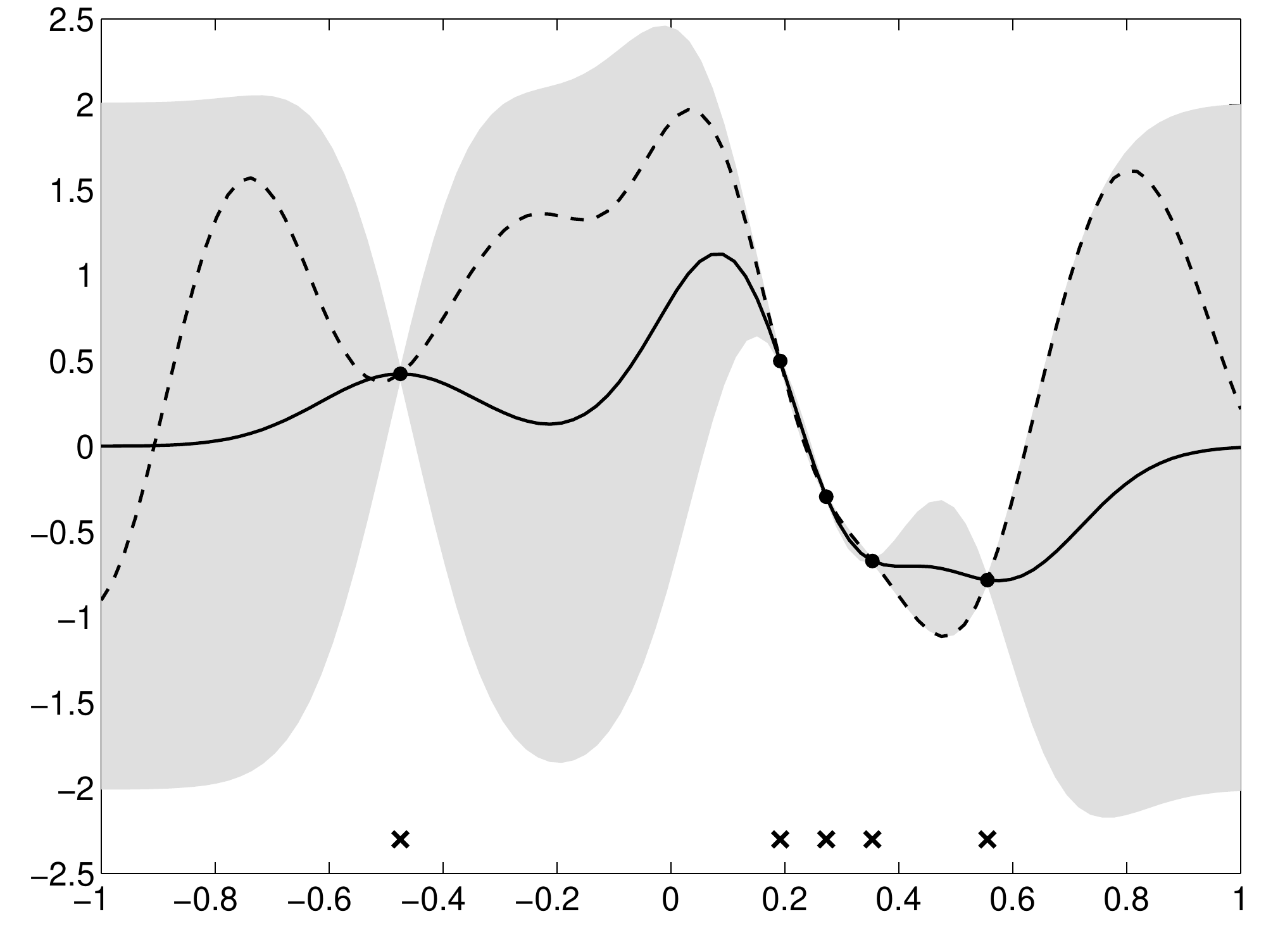}\label{figure:condLatent5}}\hfill
  \subfigure[Output one for $K=5$]{ 
  \includegraphics[width=0.30\textwidth]{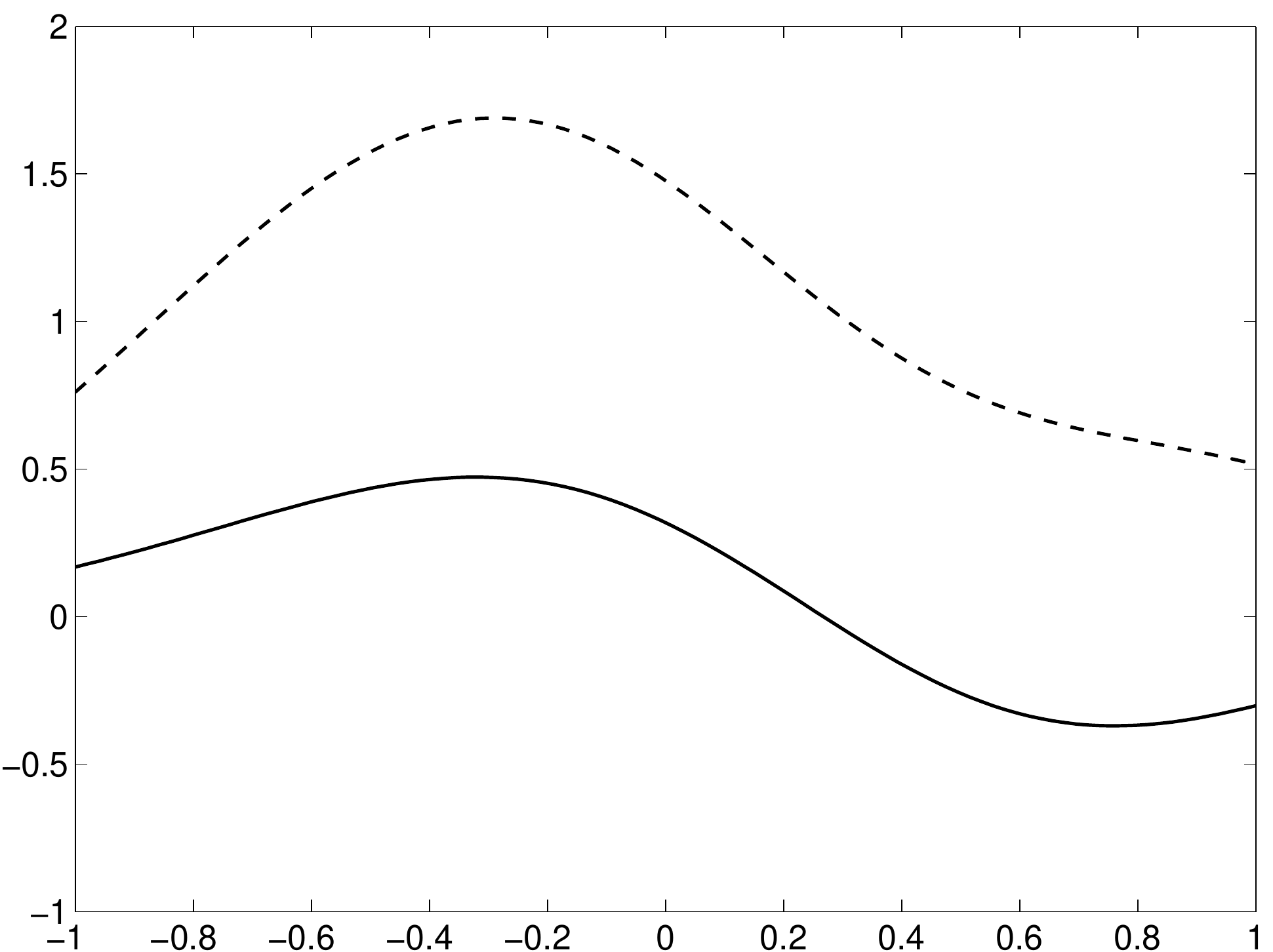}\label{figure:intrSolvedOne5}}\hfill
  \subfigure[Output two for $K=5$]{ 
  \includegraphics[width=0.30\textwidth]{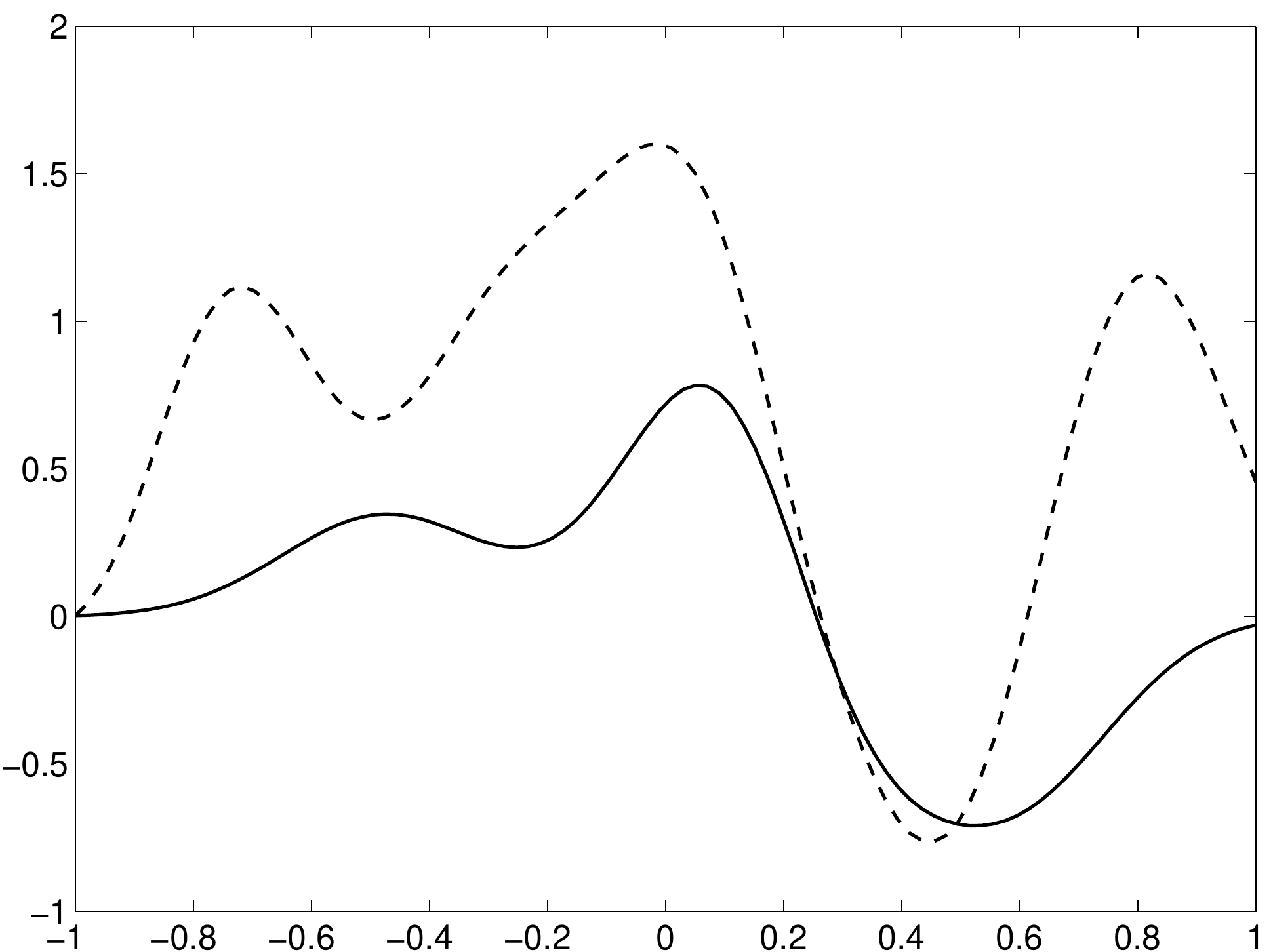}\label{figure:intrSolvedTwo5}}\\
  \subfigure[Conditional prior for $K=10$]{ 
  \includegraphics[width=0.30\textwidth]{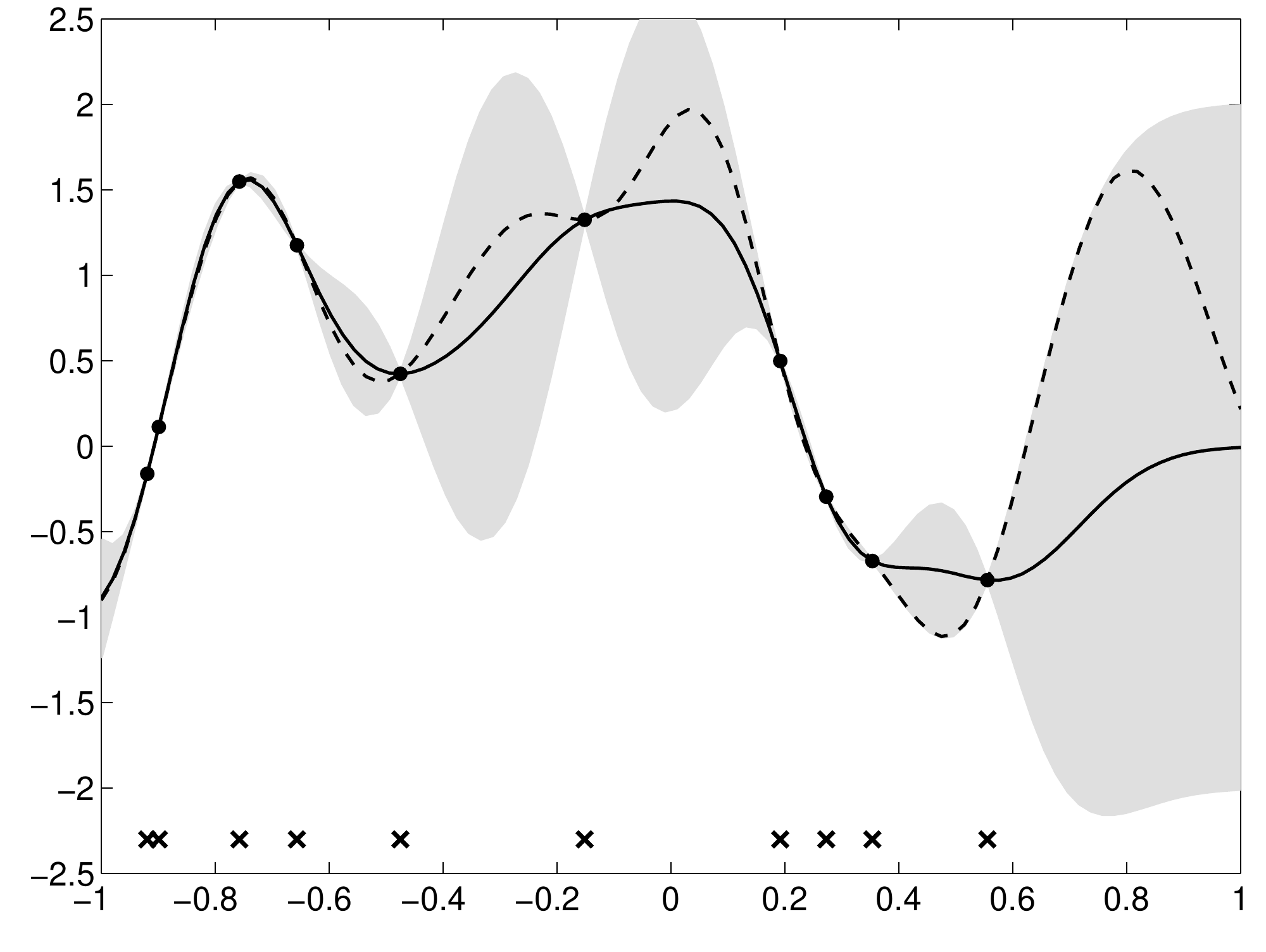}\label{figure:condLatent10}}\hfill
  \subfigure[Output one for $K=10$]{ 
  \includegraphics[width=0.30\textwidth]{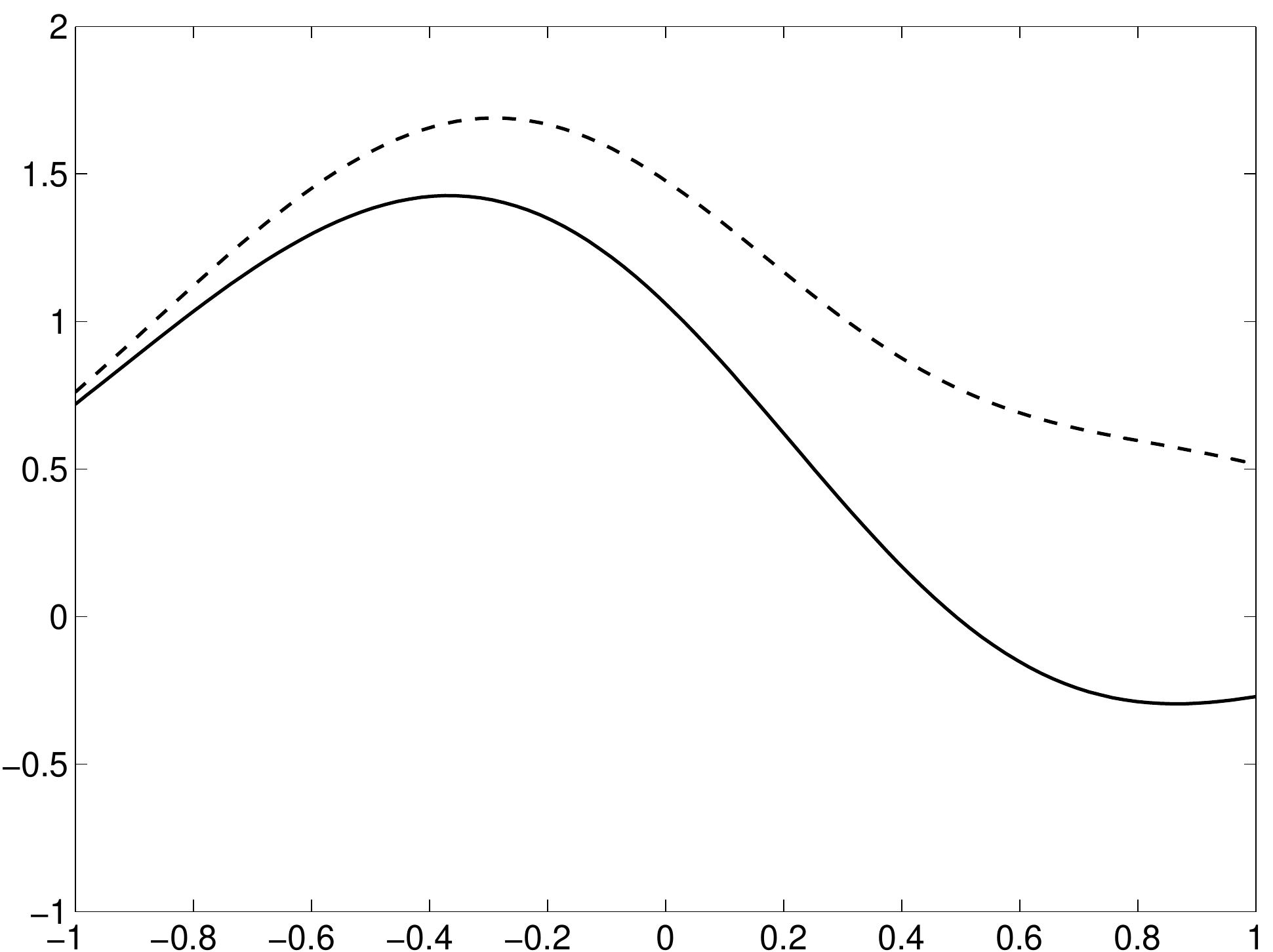}\label{figure:intrSolvedOne10}}\hfill
  \subfigure[Output two for $K=10$]{ 
  \includegraphics[width=0.30\textwidth]{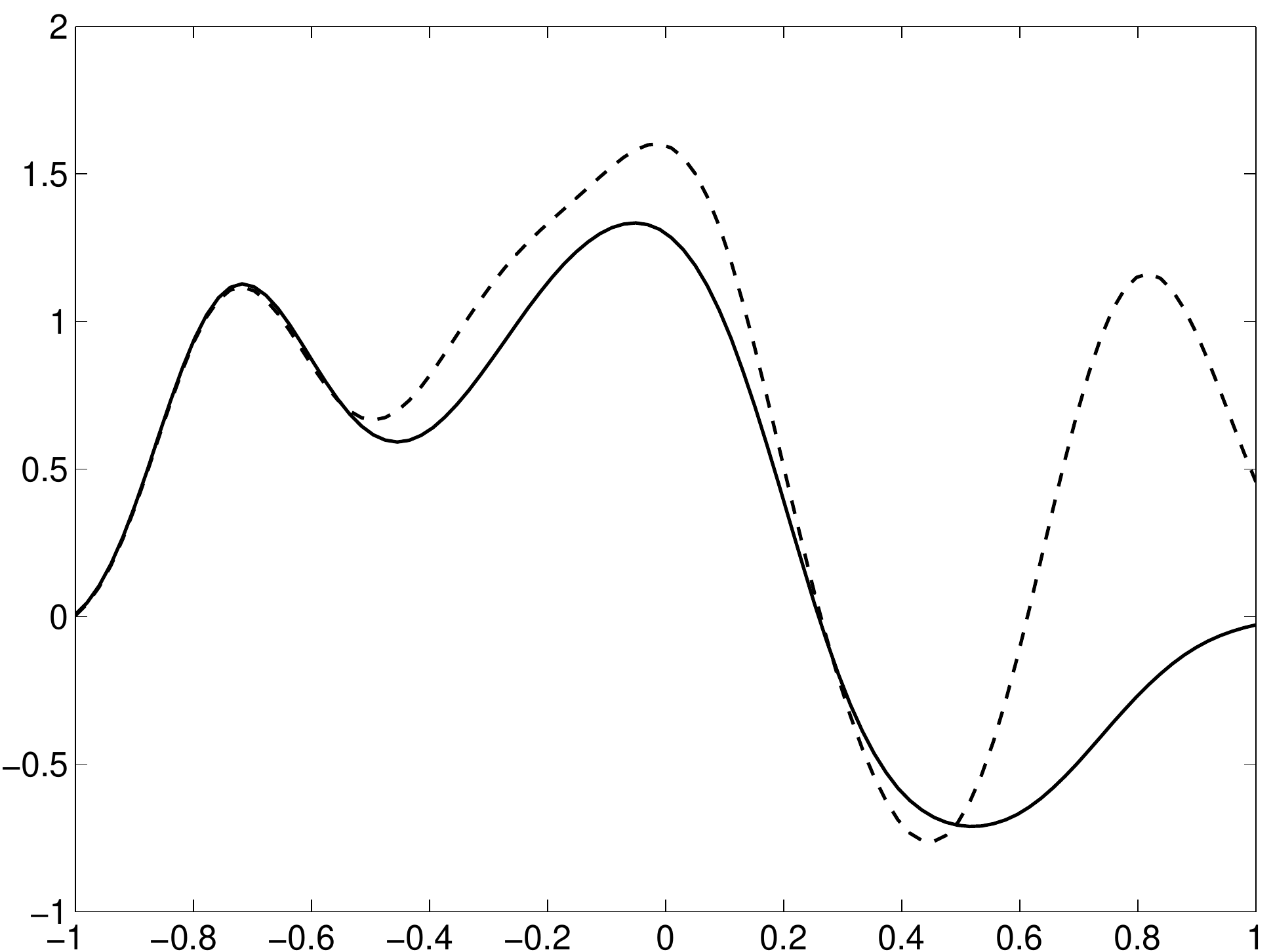}\label{figure:intrSolvedTwo10}}\\
  \subfigure[Conditional prior for $K=30$ ]{ 
  \includegraphics[width=0.30\textwidth]{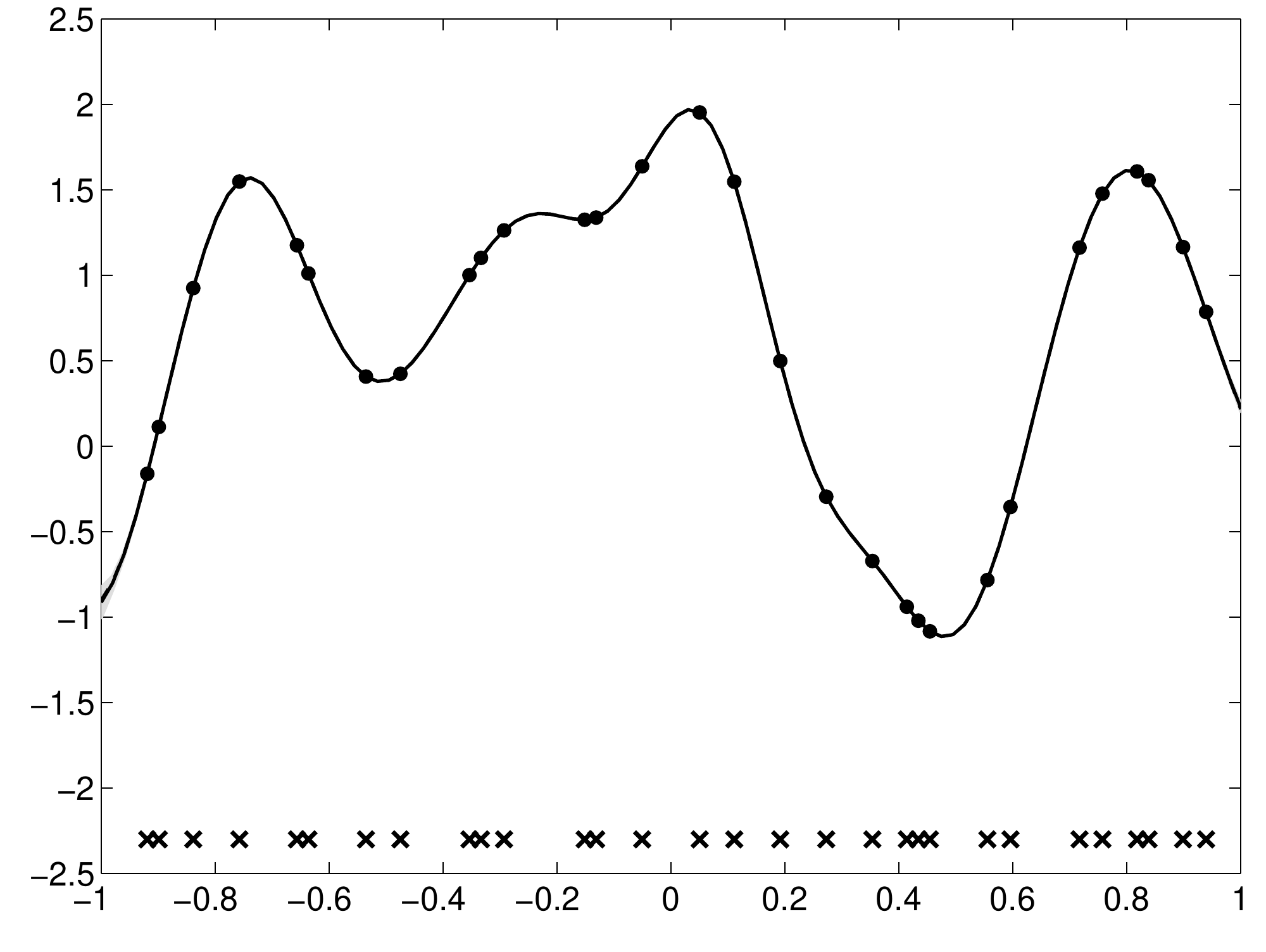}\label{figure:condLatent30}}\hfill
  \subfigure[Output one for $K=30$]{ 
  \includegraphics[width=0.30\textwidth]{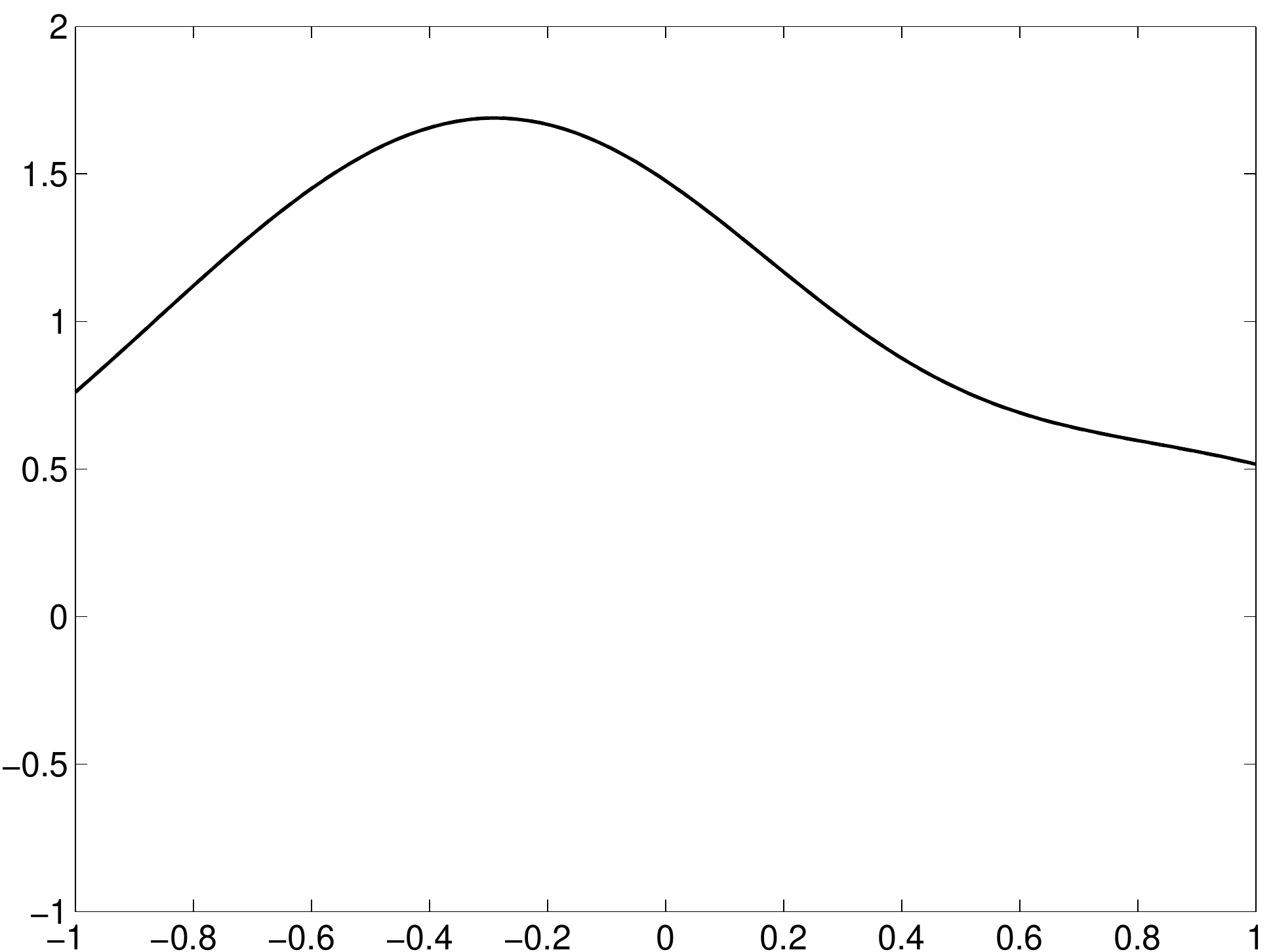}\label{figure:intrSolvedOne30}}\hfill
  \subfigure[Output two for $K=30$]{ 
  \includegraphics[width=0.30\textwidth]{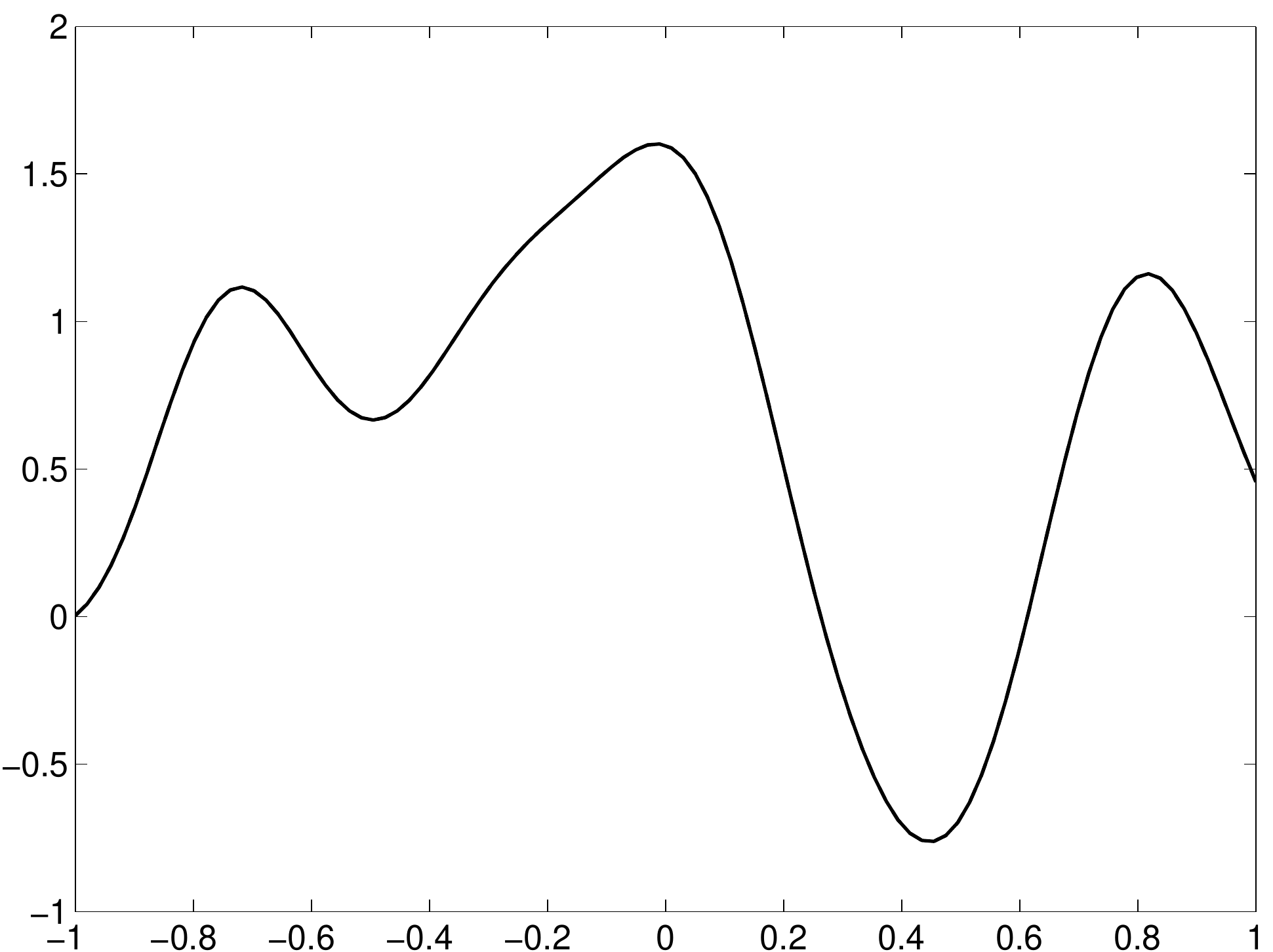}\label{figure:intrSolvedTwo30}}
  \small{\caption{ Conditional prior and two ouputs for different values of $K$. The first column, figures 
\ref{figure:condLatent5}, \ref{figure:condLatent10} and \ref{figure:condLatent30}, 
shows the mean and confidence intervals of the conditional prior distribution using one input function and two output
functions. The dashed line represents one sample from the prior. Conditioning over a few points of this sample, shown
as black dots, the conditional mean and conditional covariance are computed. The solid line represents the conditional mean 
and the shaded region corresponds to 2 standard deviations away from the mean. The second column, 
\ref{figure:intrSolvedOne5}, \ref{figure:intrSolvedOne10} and \ref{figure:intrSolvedOne30},
shows the solution to equation \eqref{eq:convolution} for output one 
using the sample from the prior (dashed line) and the conditional 
mean (solid line), for different values of $K$. The third column, 
\ref{figure:intrSolvedTwo5}, \ref{figure:intrSolvedTwo10} and \ref{figure:intrSolvedTwo30},
shows the solution to equation \eqref{eq:convolution} for output two, again for different values of $K$.} 
\label{figure:example:condPrior}}
\end{center}
\end{figure*}

For the set of approximations, instead of drawing a sample from $u(\boldz)$, we first 
draw a sample from a finite representation of $u(\boldz)$,
$\mathbf{u}(\mathbf{Z})=\left[u(\mathbf{z}_1),\ldots,u(\mathbf{z}_{K})\right]^\top$, 
where $\mathbf{Z}=\left\{\boldz_k\right\}_{k=1}^K$ is the set of input vectors at which $u(\mathbf{z})$ is evaluated. Due to
the properties of a Gaussian process, $p(\boldu(\mathbf{Z}))$ follows a multivariate Gaussian distribution. Conditioning on 
$\boldu(\mathbf{Z})$, we next sample from the conditional prior $p(u(\boldz)|\boldu(\mathbf{Z}))$ and use 
this function to solve the 
convolution integral for each $f_d(\boldx)$.\footnote{
For simplicity in the notation, we just write $\boldu$ to refer to $\boldu(\mathbf{Z})$.} 
Under this generative approach, we can approximate each function 
$f_d(\boldx)$ using
\begin{align}\label{eq:conv:approx}
f_d(\boldx) &\approx \int_{\mathcal{X}}G_{d}(\mathbf{x}-\mathbf{z})\ex \left[u(\boldz)|\boldu\right]\dif \boldz. 
\end{align}
Replacing $u(\boldz)$ for $\ex \left[u(\boldz)|\boldu\right]$ is a reasonable approximation as long as $u(\boldz)$ be
a smooth function so that the infinite dimensional object $u(\boldz)$ can be summarized by $\boldu$. Figure 
\ref{figure:example:condPrior} shows a cartoon example of the quality of the approximations for two outputs as the size of 
the set $\mathbf{Z}$ increases. The first column represents the conditional prior $p\left(u(\boldz)|\boldu\right)$ for 
a particular choice of $u(\boldz)$. The second and third columns represent the outputs $f_1(\boldx)$ and $f_2(\boldx)$ 
obtained when using equation \eqref{eq:conv:approx}.

Using expression \eqref{eq:conv:approx}, the likelihood function for $\mathbf{f}$ follows 
\begin{align*}
p(\mathbf{f}|\boldu,\mathbf{Z},\boldX,\bm{\theta})&=\mathcal{N}\left(\mathbf{K_{f,u}}\mathbf{K}^{-1}_{\mathbf{u,u}}\boldu,
\mathbf{K_{f,f}}-\mathbf{K_{f,u}}\mathbf{K}^{-1}_{\mathbf{u,u}}\mathbf{K}^{\top}_{\mathbf{f,u}}\right),
\end{align*}
where $\mathbf{K_{u,u}}$ is the covariance matrix between the samples
from the latent function $\boldu(\mathbf{Z})$, with elements given by
$k_{u,u}(\mathbf{z},\mathbf{z'})$ and $\mathbf{K_{f,u}}=\mathbf{K^\top_{u,f}}$ is the cross-covariance
matrix between the latent function $u(\mathbf{z})$ and the
outputs $f_d(\mathbf{x})$, with elements $\cov\left[f_d(\mathbf{x}),u(\mathbf{z})\right]$ in \eqref{eq:covy:yu}. 
Given the set of points $\boldu$, we can have different assumptions about the uncertainty of the outputs in the 
likelihood term. For example, we could assume that the outputs are independent or 
uncorrelated, keeping only the uncertainty involved for each output in the likelihood term.
Other approximation would assume that the outputs are deterministic, this is  
$\mathbf{K_{f,f}}=\mathbf{K_{f,u}}\mathbf{K}^{-1}_{\mathbf{u,u}}\mathbf{K}^{\top}_{\mathbf{f,u}}$. The only uncertainty left 
would be due to the prior $p(\boldu)$. Next, we present 
different approximations of the covariance of the likelihood that lead to a reduction in computational complexity.

\subsubsection{Partial Independence}
We assume that the set of outputs $\mathbf{f}$ are independent given the latent function $\mathbf{u}$, leading
to the following expression for the likelihood 
\begin{align*}
p(\mathbf{f}|\mathbf{u},\mathbf{Z},\mathbf{X},\bm{\theta})=\prod_{d=1}^Dp(\mathbf{f}_d|
\mathbf{u},\mathbf{Z},\mathbf{X},\bm{\theta})
=\prod_{d=1}^D\mathcal{N}\left(\mathbf{K}_{\mathbf{f}_d,\mathbf{u}}\mathbf{K}^{-1}_
{\mathbf{u,u}}\mathbf{u},\mathbf{K}_{\mathbf{f}_d,\mathbf{f}_d}-\mathbf{K}_{\mathbf{f}_d,\mathbf{u}}
\mathbf{K}^{-1}_{\mathbf{u,u}}\mathbf{K}_{\mathbf{u},\mathbf{f}_d}\right).
\end{align*}
We rewrite this product of multivariate Gaussians as a single
Gaussian with a block diagonal covariance matrix, including the uncertainty about the independent processes
\begin{equation}
p(\mathbf{y}|\mathbf{u},\mathbf{Z},\mathbf{X},\bm{\theta})=\mathcal{N}\left(\mathbf{K_{f,u}}\mathbf{K}^{-1}_{\mathbf{u,u}}
\mathbf{u},\mathbf{D}+\bm{\Sigma}\right)\label{eq:likelihood:sparse:pitc}
\end{equation}
where
$\mathbf{D}=\bdiag\left[\mathbf{K_{f,f}}-\mathbf{K_{f,u}}\mathbf{K}^{-1}_{\mathbf{u,u}}\mathbf{K_{u,f}}\right]$,
and we have used the notation $\bdiag\left[\mathbf{G}\right]$ to
indicate that the block associated with each output of the matrix
$\mathbf{G}$ should be retained, but all other elements should be set
to zero. We can also write this as $\mathbf{D}=
\left[\mathbf{K_{f,f}}-\mathbf{K_{f,u}}\mathbf{K}^{-1}_{\mathbf{u,u}}\mathbf{K_{u,f}}\right]\odot\mathbf{M}$
where $\odot$ is the Hadamard product and $\mathbf{M}=
\mathbf{I}_{D}\otimes\mathbf{1}_N $, $\mathbf{1}_N$ being the
$N\times N$ matrix of ones.
We now marginalize the values of the samples from the latent function by using 
its process prior, this means $p(\mathbf{u}|\mathbf{Z})=\mathcal{N}(\mathbf{0},\mathbf{K_{u,u}})$. 
This leads to the following marginal likelihood,
\begin{equation}
p(\mathbf{y}|\mathbf{Z},\mathbf{X},\bm{\theta}) =  \int
p(\mathbf{y}|\mathbf{u},\mathbf{Z},\mathbf{X},\bm{\theta})p(\mathbf{u}|\mathbf{Z})\dif{\mathbf{u}}= 
\mathcal{N}\left(\mathbf{0},\mathbf{D}+\mathbf{K_{f,u}}\mathbf{K}^{-1}_{\mathbf{u,u}}\mathbf{K_{u,f}}+
\bm{\Sigma}\right).\label{eq:marginal:sparse}
\end{equation}
Notice that, compared to \eqref{eq:marginal:full}, the full covariance
matrix $\mathbf{K_{f,f}}$ has been replaced by the low rank covariance
$\mathbf{K_{f,u}}\mathbf{K}^{-1}_{\mathbf{u,u}}\mathbf{K_{u,f}}$ in
all entries except in the diagonal blocks corresponding to
$\mathbf{K}_{\mathbf{f}_d,\mathbf{f}_{d'}}$. Depending on our choice of
$K$ the inverse of the low rank approximation to the covariance is
either dominated by a $O(DN^3)$ term or a $O(K^2DN)$ term. Storage of
the matrix is $O(N^2D)+O(NDK)$. Note that if we set $K=N$ these reduce
to $O(N^3D)$ and $O(N^2D)$ respectively. Rather neatly this matches
the computational complexity of modeling the data with $D$ independent
Gaussian processes across the outputs.

The functional form of \eqref{eq:marginal:sparse} is almost identical
to that of the PITC approximation \citep{Quinonero:unifying05}, with
the samples we retain from the latent function providing the same role
as the \emph{inducing values} in the partially independent training
conditional (PITC) approximation. This is perhaps not surprising given
that the PITC approximation is also derived by making conditional
independence assumptions. A key difference is that in PITC it is not
obvious which variables should be grouped together when making these
conditional independence assumptions, here it is clear from the
structure of the model that each of the outputs should be grouped
separately. However, the similarities are such that we find it
convenient to follow the terminology of \citet{Quinonero:unifying05}
and also refer to our approximation as a PITC approximation.

\subsubsection{Full Independence}
We can be inspired by the analogy of our approach to the PITC
approximation and consider a more radical factorization of the
likelihood term. In the fully independent training conditional (FITC)
\citep{Snelson:pseudo05,Snelson:local07} a factorization across the
data points is assumed. For us that would lead to the following
expression for conditional distribution of the output functions given
the inducing variables,
\[
p(\mathbf{f}|\mathbf{u},\mathbf{Z},\mathbf{X},\bm{\theta})=
\prod_{d=1}^D\prod_{n=1}^Np(f_{n,d}|\mathbf{u},\mathbf{Z},\mathbf{X},\bm{\theta}),
\] 
which can be briefly expressed through
\eqref{eq:likelihood:sparse:pitc} with
$\mathbf{D}=\diag\left[\mathbf{K_{f,f}}-\mathbf{K_{f,u}}\mathbf{K}^{-1}_{\mathbf{u,u}}\mathbf{K_{u,f}}\right]=
\left[\mathbf{K_{f,f}}-\mathbf{K_{f,u}}\mathbf{K}^{-1}_{\mathbf{u,u}}\mathbf{K_{u,f}}\right]\odot\mathbf{M}$,
with $\mathbf{M}= \mathbf{I}_{D}\otimes\mathbf{I}_N $. The marginal likelihood, including the uncertainty about the 
independent processes is given by equation \eqref{eq:marginal:sparse} with the diagonal form for $\mathbf{D}$, 
leading to the fully independent training conditional (FITC) approximation \citep{Snelson:pseudo05,Quinonero:unifying05}.

\subsubsection{Deterministic likelihood}
In \citet{Quinonero:unifying05} the relationship between the projected
process approximation and the FITC and PITC approximations is
elucidated. They show that if, given the set of values $\mathbf{u}$,
the ouputs are deterministic, the likelihood term of equation
\eqref{eq:conv:approx} can be simplified as
\begin{align*}
  p(\mathbf{f}|\boldu,\mathbf{Z},\boldX,\bm{\theta})&=\mathcal{N}\left(\mathbf{K_{f,u}}\mathbf{K}^{-1}_{\mathbf{u,u}}\boldu,
    \mathbf{0}\right).
\end{align*}
Marginalizing with respect to the latent function using
$p(\boldu|\mathbf{Z})=\mathcal{N}(\mathbf{0},\mathbf{K_{u,u}})$ and
including the uncertainty about the independent processes, we obtain
the marginal likelihood as
\begin{align*}
  p(\mathbf{y}|\mathbf{Z},\boldX,\bm{\theta})&=\int
  p(\mathbf{y}|\boldu,\mathbf{Z},\boldX,\bm{\theta})p(\boldu|
  \mathbf{Z})\dif \boldu=
  \mathcal{N}\left(\mathbf{0},\mathbf{K_{f,u}}\mathbf{K}^{-1}_{\mathbf{u,u}}\mathbf{K}^{\top}
    _{\mathbf{f,u}}+\bm{\Sigma}\right).
\end{align*}
In other words, we approximate the full covariance $\mathbf{K_{f,f}}$
using the low rank approximation
$\mathbf{K_{f,u}}\mathbf{K}^{-1}_{\mathbf{u,u}}\mathbf{K}^{\top}_{\mathbf{f,u}}$. Using
this new marginal likelihood to estimate the parameters $\bm{\theta}$
reduces computational complexity to $O(K^2DN)$.  The approximation
obtained has similarities with the projected latent variables (PLV)
method also known as the projected process approximation (PPA) or the
deterministic training conditional (DTC) approximation
\citep{Csato:sparse00,Seeger:fast03,Quinonero:unifying05,Rasmussen:book06}.
For this reason we refer to this approximation as the deterministic
training conditional approximation (DTC) for multiple output Gaussian
processes.

\subsubsection{Additional independence assumptions.} 

As mentioned before, we can consider different conditional
independence assumptions for the likelihood term. One further
assumption that is worth mentioning considers conditional
independencies across data points and dependence across outputs. This
would lead to the following likelihood term
\[
p(\mathbf{f}|\mathbf{u},\mathbf{Z},\mathbf{X},\bm{\theta})=
\prod_{n=1}^Np(\overline{\boldf}_{n}|\mathbf{u},\mathbf{Z},\mathbf{X},\bm{\theta}),
\] 
where $\overline{\boldf}_{n}=[f_1(\boldx_n),f_2(\boldx_n),\ldots,
f_D(\boldx_n)]^{\top}$. We can use again equation
\eqref{eq:likelihood:sparse:pitc} to express the likelihood. In this
case, if the matrix $\mathbf{D}$ is a partitioned matrix with blocks
$\mathbf{D}_{d,d'}\in\Re^{N\times N}$, each block $\mathbf{D}_{d,d'}$
would be given as $\mathbf{D}_{d,d'}=
\diag\left[\mathbf{K}_{\boldf_d,\boldf_{d'}}-\mathbf{K}_{\boldf_d,\boldu}\mathbf{K}^{-1}_{\mathbf{u,u}}\mathbf{K}_{\boldu,\boldf_{d'}}
\right]$. For cases in which $D> N$, that is, the number of outputs is
greater than the number of data points, this approximation may be more
accurate than PITC. For cases where $D< N$ it may be less accurate
than PITC, but faster to compute.\footnote{Notice that if we work with the block diagonal matrices $\mathbf{D}_{d,d'}$,
we would need to invert the full matrix $\mathbf{D}$. However, since the blocks $\mathbf{D}_{d,d'}$ 
are diagonal matrices themselves, the 
inversion can be done efficiently using, for example, a block Cholesky decomposition. Furthermore, we would be restricted
to work with isotopic input spaces. Alternatively, we could rearrange the elements of the matrix $\mathbf{D}$ so that the  
blocks of the main diagonal are the covariances associated with the vectors $\overline{\boldf}_{n}$.}

\subsection{Posterior and predictive distributions}

Combining the likelihood term for each approximation with
$p(\mathbf{u}|\mathbf{Z})$ using Bayes' theorem, the posterior
distribution over $\mathbf{u}$ is obtained as
\begin{align}\label{eq:posterior:sparse}
p(\mathbf{u}|\mathbf{y},\mathbf{X},\mathbf{Z},\bm{\theta})&=
\mathcal{N}\left(\mathbf{K_{u,u}}\mathbf{A}^{-1}\mathbf{K_{u,f}}(\mathbf{D}+\bm\Sigma)^{-1}\mathbf{y},
\mathbf{K_{u,u}}\mathbf{A}^{-1}\mathbf{K_{u,u}}\right)
\end{align}
where
$\mathbf{A}=\mathbf{K}_\mathbf{u,u}+\mathbf{K_{u,f}}(\mathbf{D}+\bm\Sigma)^{-1}\mathbf{K_{f,u}}$ and $\mathbf{D}$ follows
a particular form according to the different approximations: for PITC it equals $\mathbf{D}=\bdiag\left[
\mathbf{K_{f,f}}-\mathbf{K_{f,u}}\mathbf{K}^{-1}_{\mathbf{u,u}}\mathbf{K_{u,f}}\right]$, for FITC $\mathbf{D}=
\diag\left[\mathbf{K_{f,f}}-\mathbf{K_{f,u}}\mathbf{K}^{-1}_{\mathbf{u,u}}\mathbf{K_{u,f}}\right]$ and for DTC 
$\mathbf{D}=\mathbf{0}$.
The predictive distribution is expressed through the integration of the likelihood term, evaluated at 
$\mathbf{X_*}$, with \eqref{eq:posterior:sparse}, giving
\begin{align}
p(\mathbf{y}_*|\mathbf{y},\mathbf{X},\mathbf{X}_*,\mathbf{Z},\bm{\theta})=&\int
p(\mathbf{y}_*|\mathbf{u},\mathbf{Z},\mathbf{X}_*,\bm{\theta})p(\mathbf{u}|\mathbf{y},\mathbf{X},
\mathbf{Z},\bm{\theta})\dif{\mathbf{u}}
\nonumber\\
=&\mathcal{N}\left(\mathbf{K_{f_*,u}}\mathbf{A}^{-1}\mathbf{K_{u,f}}(\mathbf{D}+\bm\Sigma)^{-1}\mathbf{y},\mathbf{D_*}+
\mathbf{K_{f_*,u}}\mathbf{A}^{-1}\mathbf{K_{u,f_*}}+\bm\Sigma\right),
\nonumber
\end{align}
with
$\mathbf{D_*}=\bdiag\left[\mathbf{K_{f_*,f_*}}-\mathbf{K_{f_*,u}}\mathbf{K}^{-1}_{\mathbf{u,u}}\mathbf{K_{u,f_*}}\right]$ 
for PITC, $\mathbf{D_*}=\diag\left[\mathbf{K_{f_*,f_*}}-\mathbf{K_{f_*,u}}\mathbf{K}^{-1}_{\mathbf{u,u}}
\mathbf{K_{u,f_*}}\right]$ for FITC and $\mathbf{D_*}=\mathbf{0}$ for DTC.

\subsection{Fitting the Model}

The marginal likelihood approximation for the PITC, FITC and DTC variants
for sparse multioutput Gaussian processes is a function of both the
parameters of the covariance function and the location of the inputs
for the inducing variables. One of the key ideas presented in
\citet{Snelson:pseudo05} was that we should optimize with respect to
the location of these inducing variables. Previously, the inducing
variables were taken to be a subset of the data variables
\citep{Csato:sparse00,Williams:nystrom00}. However, a method of
choosing which subset of the data is required
\citep{Smola:sparsegp00,Seeger:fast03}. Such criteria can be expensive to compute
and also lead to fluctuations in the approximation to the
log-likelihood when the subset changes. This causes problems as while
the parameters of the Gaussian process are optimized, the optimal
subset of the inducing inputs will also change. Convergence is
therefore difficult to monitor. The key advantage of optimizing the
inducing input locations, $\mathbf{Z}$, across the entire input domain is that
convergence of the likelihood will be smooth. In appendix \ref{subsection:derivatives:PITC:FITC} we
include the derivatives of the marginal likelihood wrt the matrices
$\mathbf{K_{f,f}},\mathbf{K_{u,f}}$ and $\mathbf{K_{u,u}}$.

\section{Experimental evaluation}\label{section:experiments}

In this section we present results of applying the sparse methods in pollutant metal prediction, exam score prediction and
the prediction of transcription factor behavior in a gene-network. First, though, we ilustrate the performance of the sparse
method in a toy example.\footnote{Code to run all simulations in this section is available at 
\url{http://www.cs.manchester.ac.uk/~neill/multigp/}}

\subsection{A toy example}\label{section:toy1D}

For the toy experiment, we employ the kernels constructed in Example 1 of section \ref{section:examples:kernels}. 
The toy problem consists of $D=4$ outputs, one
latent function, $Q=1$ and one input dimension. The training data was sampled from the full GP with the
following parameters, $S_{1,1}=S_{2,1}=1$, $S_{3,1}=S_{4,1}=5$,
$P_{1,1}=P_{2,1}=50$, $P_{3,1}=300,P_{4,1}=200$ for the outputs and
$\Lambda_{1}=100$ for the latent function. For the independent processes,
$w_d\left(\mathbf{x}\right)$, we simply added white noise separately
to each output so we have variances $\sigma_1^2=\sigma_2^2=0.0125$,
$\sigma_3^2=1.2$ and $\sigma_4^2=1$. We generate $N=500$ observation points for each
output and use $200$ observation points (per output) for training the full and the sparse multiple output GP 
and the remaining $300$ observation points for testing. We repeated the same experiment setup $10$ times and compute 
the standardized 
mean square error (SMSE)  and the mean standardized log loss (MSLL) as defined
in \citet{Rasmussen:book06}. For the sparse methods we use $K=30$ inducing inputs. We sought the kernel parameters and
the positions of the inducing inputs through maximizing the marginal likelihood using a scaled conjugate gradient 
algorithm. Initially the inducing inputs are equally spaced between the interval $[-1,1]$. 

\begin{figure}[ht!]
\begin{center}
\subfigure[$y_4(x)$ using the full GP]{ \label{fig:out4:full:train:test}
\resizebox{0.47\textwidth}{!}{\includegraphics{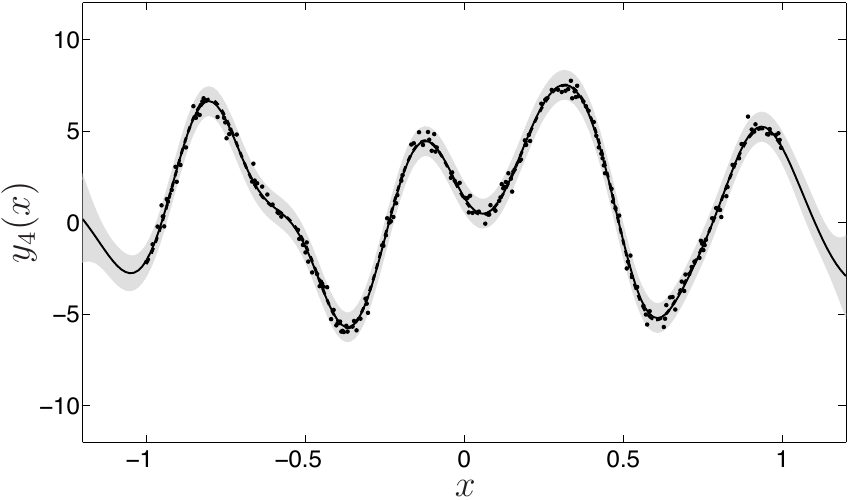}}}
\subfigure[$y_4(x)$ using the DTC approximation]{ \label{fig:out4:dtc:train:test}
\resizebox{0.47\textwidth}{!}{\includegraphics{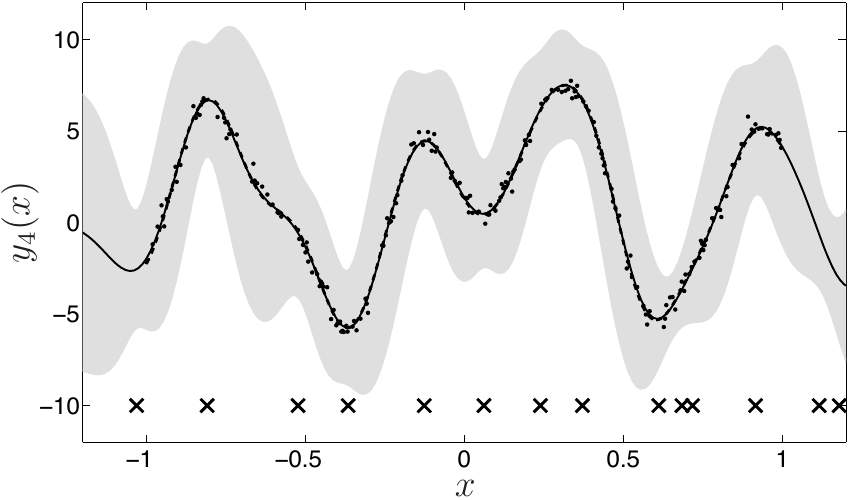}}}
\subfigure[$y_4(x)$ using the FITC approximation]{\label{fig:out4:fitc:train:test}
\resizebox{0.47\textwidth}{!}{\includegraphics{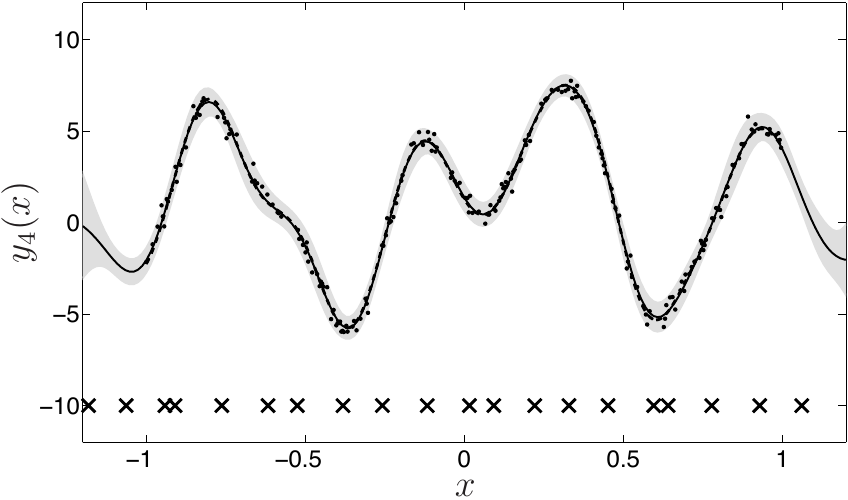}}}
\subfigure[$y_4(x)$ using the PITC approximation]{\label{fig:out4:pitc:train:test}
\resizebox{0.47\textwidth}{!}{\includegraphics{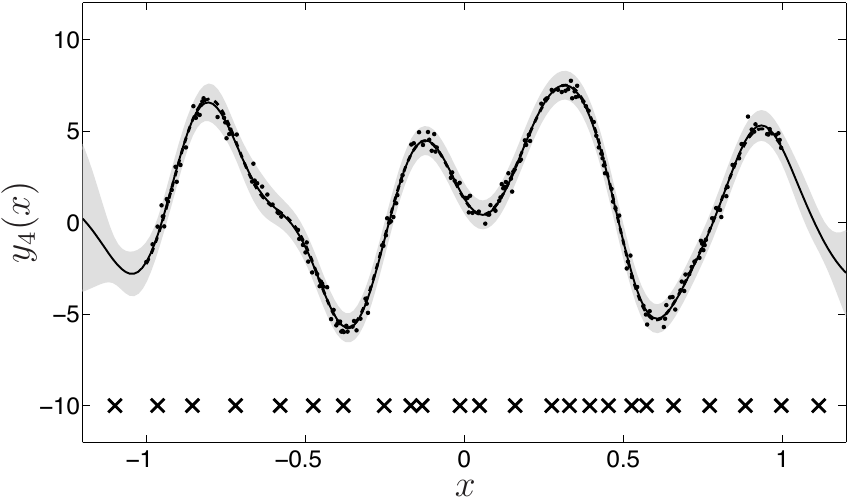}}}
\caption{Predictive mean and variance using the full multi-output GP
  and the sparse approximations for output 4. The solid line
  corresponds to the predictive mean, the shaded region corresponds to
  $2$ standard deviations of the prediction. The dashed line
  corresponds to the ground truth signal, that is, the sample from the
  full GP model without noise. In these plots the predictive mean overlaps almost exactly with the ground truth.  
  The dots are the noisy training points. The crosses
  in figures \ref{fig:out4:dtc:train:test},
  \ref{fig:out4:fitc:train:test} and \ref{fig:out4:pitc:train:test}
  correspond to the locations of the inducing inputs after
  convergence.} \label{fig:toy:train:test}
\end{center}
\end{figure}

Figure \ref{fig:toy:train:test} shows the training result of one of the ten repetitions.
The predictions shown correspond to the full GP
(Figure~\ref{fig:out4:full:train:test}), the DTC approximation (Figure~\ref{fig:out4:dtc:train:test}), 
the FITC approximation (Figure~\ref{fig:out4:fitc:train:test}) and the PITC approximation
(Figure~\ref{fig:out4:pitc:train:test}).

Tables~\ref{table:toy:smse} and \ref{table:toy:snpl} show the average prediction results over the
test set. Table~\ref{table:toy:smse}, shows that the
SMSE of the sparse approximations is similar to the one obtained with
the full GP. However, there are important differences in the values of the MSLL shown in table \ref{table:toy:snpl}.
DTC offers the worst performance. It gets better for FITC and PITC since they offer a more precise approximation to the 
full covariance. 

\begin{table}[ht!]
\begin{center}
\begin{tabular}{|c|c|c|c|c|c|}\hline
\multicolumn{1}{|c|}{Method}&\multicolumn{1}{|c|}{SMSE ${y_1(x)}$}&\multicolumn{1}{|c|}{SMSE ${y_2(x)}$}&
\multicolumn{1}{|c|}{SMSE ${y_3(x)}$}&\multicolumn{1}{|c|}{SMSE ${y_4(x)}$}\\\hline
Full GP & $1.06\pm0.08$ & $0.99\pm0.06$ & $1.10\pm0.09$ & $1.05\pm0.09$ \\
DTC & $1.06\pm0.08$ & $0.99\pm0.06$ & $1.12\pm0.09$ & $1.05\pm0.09$\\
FITC & $1.06\pm0.08$ & $0.99\pm0.06$ & $1.10\pm0.08$ & $1.05\pm0.08$ \\
PITC & $1.06\pm0.08$ & $0.99\pm0.06$ & $1.10\pm0.09$ & $1.05\pm0.09$
\\\hline
\end{tabular}
\end{center}
\caption{Standarized mean square error (SMSE) for the toy problem
over the test set. All numbers are to be multiplied by
$10^{-2}$. The experiment was repeated ten times. Table includes the
value of one standard deviation over the ten repetitions.
}\label{table:toy:smse}
\end{table}
                               
\begin{table}[ht!]
\begin{center}
\begin{tabular}{|c|c|c|c|c|c|}\hline
\multicolumn{1}{|c|}{Method}&\multicolumn{1}{|c|}{MSLL ${y_1(x)}$}&\multicolumn{1}{|c|}{MSLL ${y_2(x)}$}&
\multicolumn{1}{|c|}{MSLL ${y_3(x)}$}&\multicolumn{1}{|c|}{MSLL ${y_4(x)}$}\\\hline
Full GP & $-2.27\pm0.04$ & $-2.30\pm0.03$ & $-2.25\pm0.04$ & $-2.27\pm0.05$ \\
DTC & $-0.98\pm0.18$ & $-0.98\pm0.18$ & $-1.25\pm0.16$ & $-1.25\pm0.16$\\
FITC & $-2.26\pm0.04$ & $-2.29\pm0.03$ & $-2.16\pm0.04$ & $-2.23\pm0.05$\\
PITC & $-2.27\pm 0.04$ & $-2.30\pm0.03$ & $-2.23\pm0.04$ & $-2.26\pm0.05$
\\\hline
\end{tabular}
\end{center}
\caption{Mean standardized log loss (MSLL) for the toy problem
  over the test set. More negative values of MSLL indicate better models. 
  The experiment was repeated ten times. Table includes the
  value of one standard deviation over the ten repetitions. 
}\label{table:toy:snpl}
\end{table}

Also, the training times for iteration of each model are
$1.97\pm0.02$ secs for the full GP, $0.20\pm0.01$ secs for DTC, $0.41\pm 0.03$ for FITC
and $0.59\pm0.05$ for the PITC.

\begin{figure}[ht!]
\begin{center}
\subfigure[$y_4(x)$ using the full GP]{ \label{fig:out4:full:missing}
\resizebox{0.48\textwidth}{!}{\includegraphics{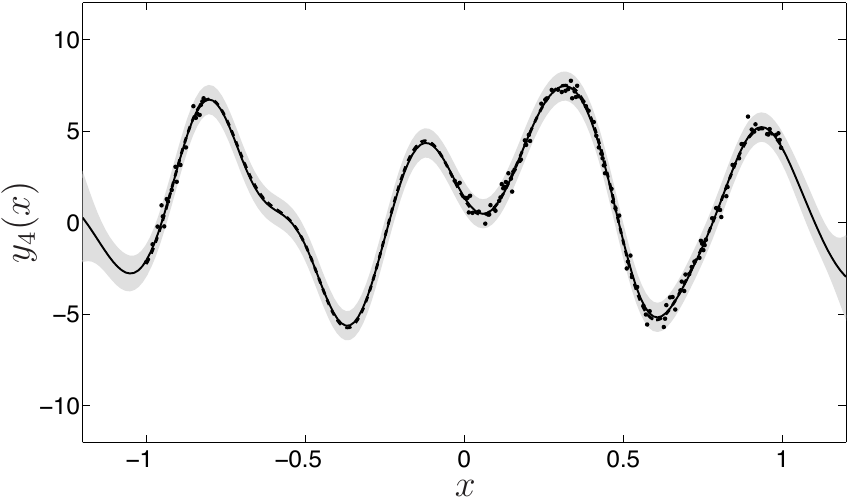}}}
\subfigure[$y_4(x)$ using an independent GP]{ \label{fig:out4:ind:missing}
\resizebox{0.48\textwidth}{!}{\includegraphics{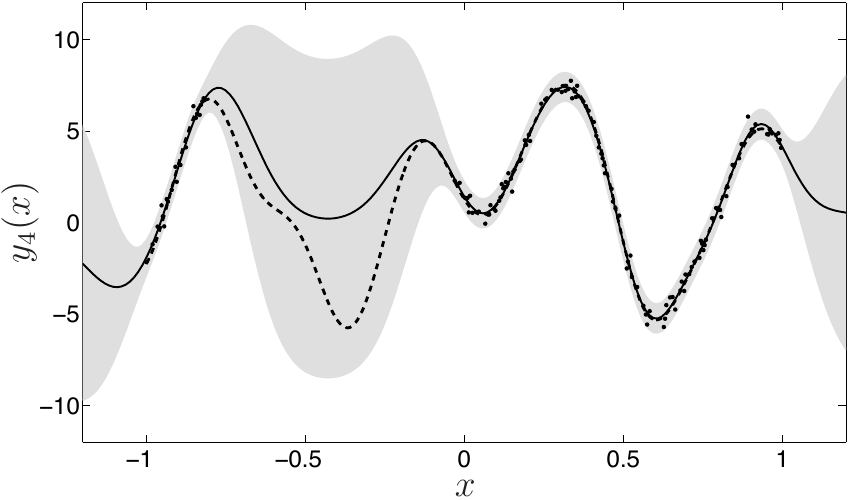}}}
\subfigure[$y_4(x)$ using the DTC
approximation]{\label{fig:out4:dtc:missing}
\resizebox{0.48\textwidth}{!}{\includegraphics{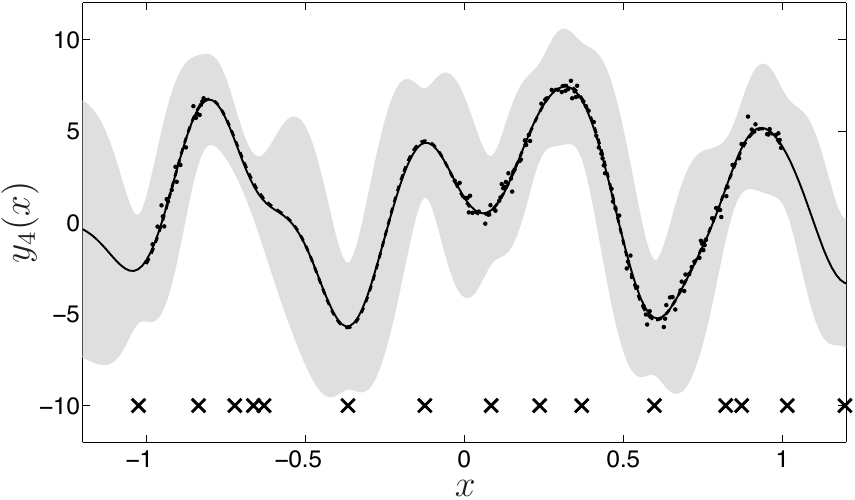}}}
\subfigure[$y_4(x)$ using the FITC
approximation]{\label{fig:out4:fitc:missing}
\resizebox{0.48\textwidth}{!}{\includegraphics{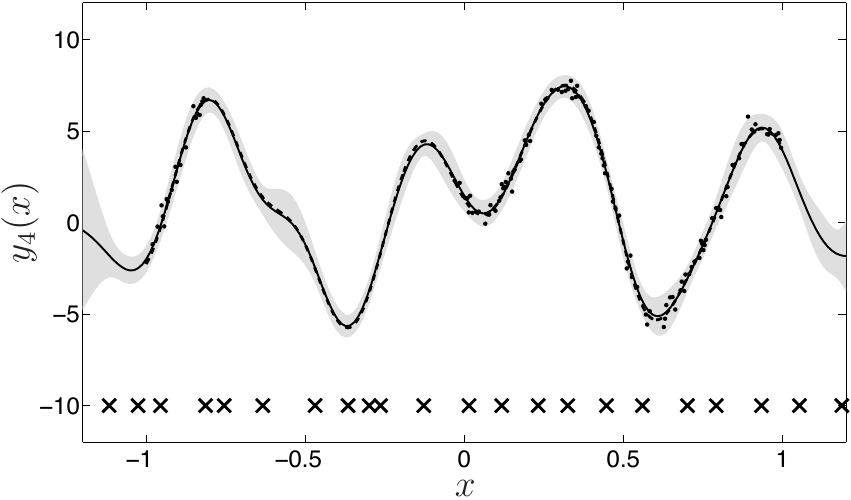}}}
\subfigure[$y_4(x)$ using the PITC
approximation]{\label{fig:out4:pitc:missing}
\resizebox{0.48\textwidth}{!}{\includegraphics{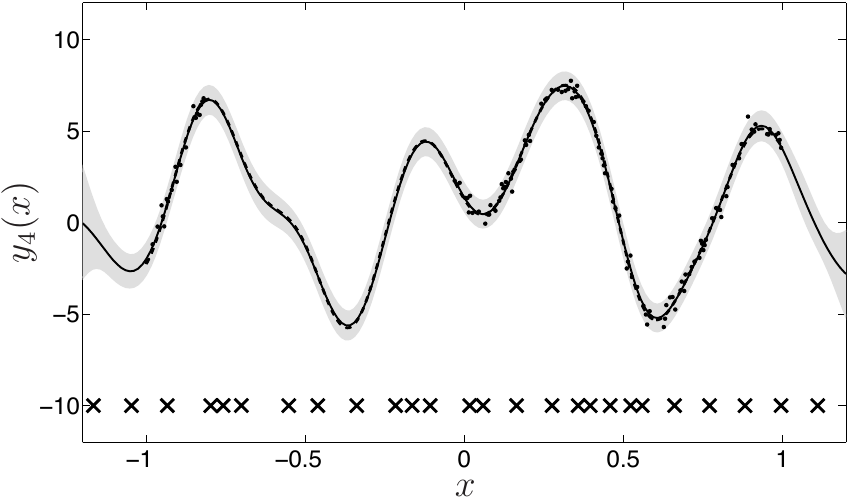}}}
\caption{Predictive mean and variance using the full multi-output
GP, the sparse approximations and an independent GP for output 4 with a range of missing observations
in the interval $[-0.8, 0.0]$. The
solid line corresponds to the mean predictive, the shaded region
corresponds to $2$ standard deviations away from the mean and the
dash line is the actual value of the signal without noise. The dots
are the noisy training points. The crosses in figures \ref{fig:out4:dtc:missing},
\ref{fig:out4:fitc:missing} and \ref{fig:out4:pitc:missing} correspond to the
locations of the inducing inputs after convergence.} \label{fig:toy:missing}
\end{center}
\end{figure}

As we have mentioned before, one important feature of multiple output prediction is that
we can exploit correlations between outputs to predict missing observations. We used a simple example 
to illustrate this point. We removed a portion of one output between $[-0.8, 0]$ from the training data in the experiment
before (as shown in Figure~\ref{fig:toy:missing}) and train the different models to predict the behavior of $y_4(x)$ for
the missing information. The predictions shown correspond to the full GP
(Figure~\ref{fig:out4:full:missing}), an independent GP
(Figure~\ref{fig:out4:ind:missing}), the DTC approximation (Figure~\ref{fig:out4:dtc:missing}), the FITC approximation
(Figure~\ref{fig:out4:fitc:missing}) and the PITC approximation
(Figure~\ref{fig:out4:pitc:missing}). The training of the sparse methods is done in the same way than in the 
experiment before.

Due to the strong dependencies between the signals, our model is
able to capture the correlations and predicts accurately the missing
information.

\subsection{Heavy Metals in the Swiss Jura}\label{section:Jura}

\begin{figure}[ht!]
\begin{center}
\includegraphics[width=1\textwidth]{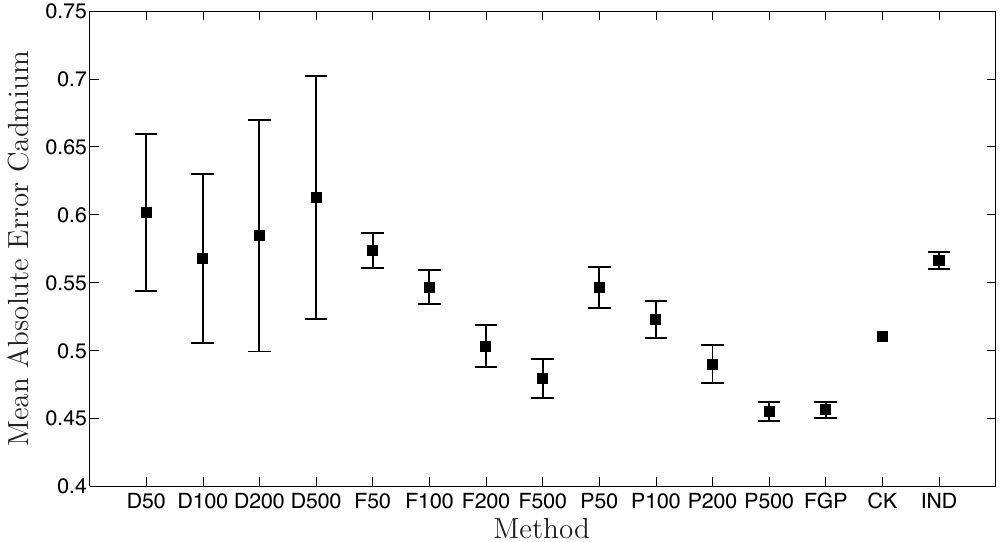}
\caption{Mean absolute error and standard deviation for prediction of
  the pollutant metal Cadmium.  The experiment was repeated ten
  times. In the bottom of the figure D$K$, F$K$, P$K$ stands for DTC,
  FITC and PITC with $K$ inducing values, respectively, FGP stands for
  full GP, CK stands for ordinary cokriging using the linear model of
  coregionalization (see \citet{Goovaerts:book97} for detailed
  description of the ordinary cokriging estimator) and IND stands for
  independent GP.} \label{fig:juraCd}
\end{center}
\end{figure}

The first example with real data that we consider is the prediction of
the concentration of several metal pollulants in a region of the Swiss
Jura. The data consist of measurements of concentrations of several
heavy metals collected in the topsoil of a $14.5$ $\mbox{km}^2$ region
of the Swiss Jura. The data is divided into a prediction set ($259$
locations) and a validation set ($100$ locations).\footnote{This data
  is available at \url{http://www.ai-geostats.org/}.} In a typical
situation, referred to as undersampled or heterotopic case, a few
expensive measurements of the attribute of interest are supplemented
by more abundant data on correlated attributes that are cheaper to
sample.  We follow the experiment described in \citet[p. 248,
249]{Goovaerts:book97} in which a \emph{primary variable} (cadmium) at
prediction locations in conjunction with some \emph{secondary
  variables} (nickel and zinc) at prediction and validation locations,
are employed to predict the concentration of the primary variable at
validation locations. We compare results of independent GP, the
different approximations described before, the full GP and ordinary
cokriging.\footnote{Cokriging is the generalization of kriging to
  multiple outputs. Within cokriging there are several alternatives,
  including simple and ordinary cokriging. Interested readers are
  referred to \citep[ch. 6]{Goovaerts:book97} for details. In the
  geostatistics literature, the usual procedure is to use the linear
  model of coregionalization to construct a valid covariance function
  and then use the cokriging estimator for making predictions.
} 
For the convolved GPs, we use one ($Q=1$) latent
function. For the sparse approximations results, a \emph{k-means}
procedure is employed first to find the initial locations of the
inducing values and then these locations are optimized in the same
optimization procedure used for the parameters. Each experiment is
repeated ten times. The result for ordinary cokriging was obtained
from \citet[p. 248, 249][]{Goovaerts:book97}. In this case, no values for
standard deviation are reported. Figure \ref{fig:juraCd} shows the results
of prediction for cadmium
(Cd). 
It can be noticed that as more inducing values are included, the
approximations follow the performance of the full GP, as would be
expected. For this particular dataset, FITC and PITC exhibit lower
variances compared to DTC. In terms of the performance, it can be seen
that PITC outperforms FITC and DTC when compared in terms of the
number of inducing points. FITC and PITC also outperform the
independent GP method, and for $200$ and $500$ inducing points, they
outperform the cokriging method.

\subsection{Exam score prediction}\label{section:exam:score}
In the second experiment with real data the goal is to predict the exam score obtained by a particular 
student belonging to a particular school. The data comes from the Inner London Education
Authority (ILEA).\footnote{This data is available at
\url{http://www.cmm.bristol.ac.uk/learning-training/multilevel-m-support/datasets.shtml}.}
It consists of examination records from 139 secondary schools in years 1985, 1986 and 1987. It is a random $50\%$ sample
with 15362 students. The input space consists of four features related to each student (year in which each student took 
the exam, gender, VR band and ethnic group) and four features related to each school (percentage of students eligible for 
free school meals, percentage of students in VR band one, school gender and school denomination). From the multiple output 
point of view, each school represents one output and the exam score of each student a particular instantiation of that 
output or $D=139$. 

We follow the same preprocessing steps employed in
\citet{Bonilla:multi07}. The only features used are the
student-dependent ones, which are categorial variables. Each of them
is transformed to a binary representation. For example, the possible
values that the variable year of the exam can take are 1985, 1986 or
1987 and are represented as $100$, $010$ or $001$. The transformation
is also applied to the variables gender (two binary variables), VR
band (four binary variables) and ethnic group (eleven binary
variables), ending up with an input space with $20$ dimensions. The
categorial nature of the data restricts the input space to $N=202$ unique
input feature vectors. However, two students represented by the same
input vector $\boldx$, and belonging both to the same school, $d$, can
obtain different exam scores. To reduce this noise in the data, we
take the mean of the observations that, within a school, share the
same input vector and use a simple heteroskedastic noise model in
which the variance for each of these means is divided by the number of
observations used to compute it.\footnote{Different noise models can
  be used. However, we employed this one so that we can compare directly to the
  results presented in \citet{Bonilla:multi07}.} For the convolved
GPs, we use one ($Q=1$) latent function.
The performace measure employed is the percentage of explained variance
defined as the total variance of the data minus the sum-squared error
on the test set as a percentage of the total data variance. It can be
seen as the percentage version of the coefficient of determination
between the test targets and the predictions.  The performace measure
is computed for ten repetitions with $75\%$ of the data in the
training set and $25\%$ of the data in the testing set.

\begin{figure}[ht!]
  \begin{center}
    \includegraphics[width=1\textwidth]{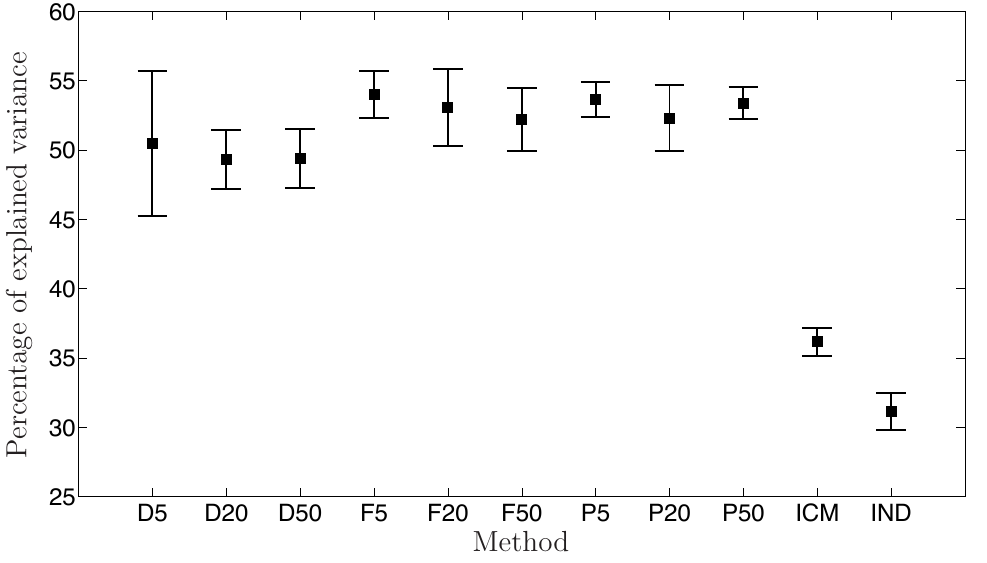}
    \caption{Mean and standard deviation of the percentage of explained
      variance for exam score prediction results on the ILEA
      dataset. The experiment was repeated ten times. In the bottom of
      the figure D$K$, F$K$, P$K$ stands for DTC, FITC and PITC with
      $K$ inducing values, respectively, ICM stands for intrinsic
      coregionalization model and IND stands for independent GPs. The
      ICM and the independent GPs results were obtained from
      \citet{Bonilla:multi07}.} \label{fig:results:school}
  \end{center}
\end{figure}

As in the Jura dataset experiment, the initial positions of the
inducing points are selected using the \emph{k-means} algorithm with
the data points as inputs to the algorithm. The positions of these
points are optimized in a scaled conjugate gradient procedure together
with the parameters of the model.  Figure \ref{fig:results:school}
shows results of the sparse methods, the ICM model and independent
GPs. The results for the ICM model are the best results presented in
\citet{Bonilla:multi07}. The independent GPs result was also obtained
from \citet{Bonilla:multi07}. It can be seen that the sparse convolved
multiple output GP framework outperforms the ICM model and the
independent GPs, even with as few as 5 inducing points. FITC and PITC
slightly outperform the DTC method, which also has greater
variances. This dataset was also employed to evaluate the performance
of the multitask kernels in \citet{Evgeniou:multitaskSVM04}. The best
result presented in this work was $34.37 \pm 0.3$.  However, due to
the averaging of the observations that we employed here, it is not
fair to compare directly against those results.

\subsection{Transcription factor regulation in the cell cycle of Yeast}\label{section:Yeast}

We now consider an application of the multiple output convolutional
model in transcriptional regulation. Microarray studies have made the
simultaneous measurement of mRNA from thousands of genes
practical. Transcription is governed by the presence of absence of
transcription factor proteins that act as switches to turn on and off
the expression of the genes. The active concentration of these
transcription factors is typically much more difficult to measure.
Several alternative methods have been proposed to infer these
activities using gene expression data and information about the
network architecture.  However, most of these methods are based on
assuming that there is an instantaneous linear relationship between
the gene expression and the protein concentration. This simplifying
assumption allows these methods to be applied on a genome wide
scale. However, it is possible to obtain a more detailed description
of the dynamics of this interaction using more realistic models that
employ differential equations. One example of this type of modeling
was presented in \citet{Barenco:ranked06}. \citet{Barenco:ranked06}
used an ordinary first order differential equation to model the
interaction between a single transcription factor and a number of
genes in a biological network motif known as a single input module. A
typical dataset of this type consists of $N$ measurements of the mRNA
abundance level of $D$ genes. The expression level $f_d(t)$ of gene
$d$ at time $t$ is related with the transcription factor protein
$u(t)$ through
\begin{align*}
  \frac{\dif f_d}{\dif t}=B_d+S_du(t)-\gamma_df_d(t),
\end{align*}
where $B_d$ is the basal transcription rate of gene $d$, $S_d$ is the
sensitivity of gene $d$ to the transcription factor and $\gamma_d$ is
the decay rate of mRNA. Solution for $f_d(t)$ was given in equation
\eqref{eq:conv:ode1} with $Q=1$ and taking into account the additional
parameter $B_d$, it follows
\begin{align*}
  f_d(t)&=\frac{B_d}{\gamma_d}+
  S_{d}\int_{0}^t\exp(-\gamma_d(t-\tau))u(\tau)\dif \tau.
\end{align*}
Given some training data for $f_d(t)$, the usual way to estimate the
dynamics of $u(t)$ is to establish an error function and minimize it
with respect to each value of $u(t)$ and the parameters of the
differential equation.

An alternative way to deal with these differential equations was
proposed by \citet{Lawrence:gpsim2007a,Gao:latent08}.  Instead of
finding a point estimate for $u(t)$, these authors proposed to put a
Gaussian process prior over $u(t)$ and use Bayesian analysis to infer
the posterior distribution for $u(t)$ using the data for
$f_d(t)$. This corresponds exactly to the multiple output convolved
Gaussian process framework that we have described in Section
\ref{section:convolution}. In this case, the latent functions $u_q(t)$
correspond to the transcription factor proteins and the outputs
$f_d(t)$ represent the gene expression data.  Due to the computational
complexity issue that we have already discussed,
\citet{Lawrence:gpsim2007a,Gao:latent08} only dealt with a reduced
number of genes. The first order differential
equation model is obviously an oversimplification of transcriptional
regulation. However, it considers more aspects of the system than
clustering, or factor analysis. The sparse approach allows us to apply
this richer model on a genome wide scale. Also, because this is done
within the framework of Gaussian processes, it is always possible to
add other, perhaps independent, terms to the covariance function to
deal with any model mismatch.
    
\begin{figure}[ht!]
\begin{center}
  \subfigure[Gene expression profile for ACE2.]{
    {\includegraphics[width=0.48\textwidth]{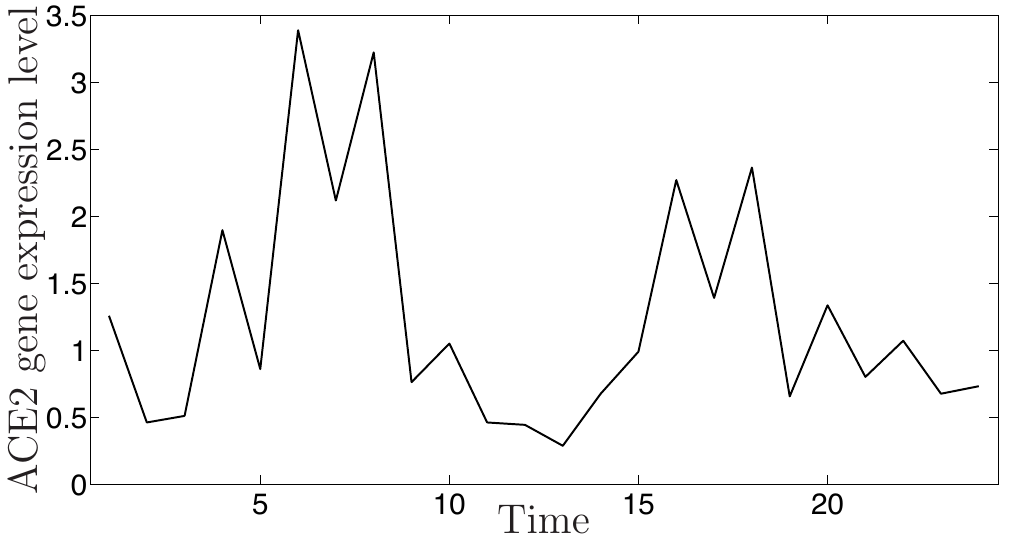}}\label{fig:ace2:gene:expression}}
  \subfigure[Inferred protein concentration for ACE2.]{
    {\includegraphics[width=0.48\textwidth]{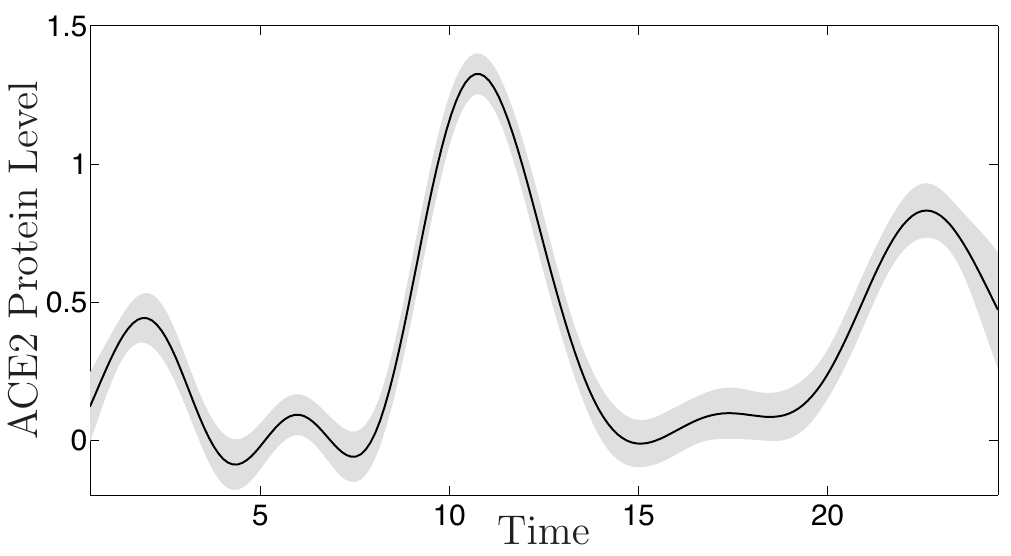}}\label{fig:ace2:factor:protein}}
  \subfigure[Gene expression profile for SWI5.]{
    {\includegraphics[width=0.48\textwidth]{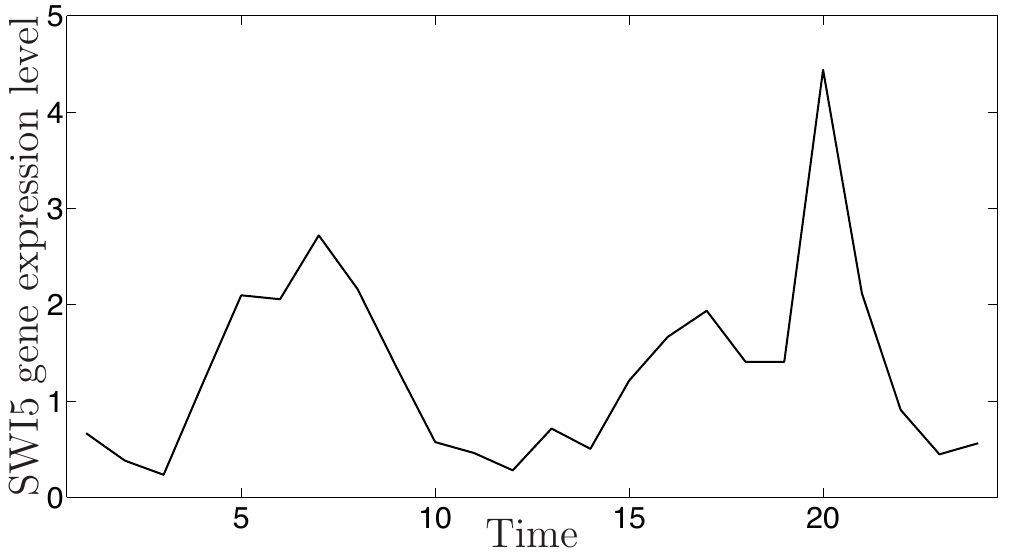}}\label{fig:swi5:gene:expression}}
  \subfigure[Inferred protein concentration for SWI5.]{
    {\includegraphics[width=0.48\textwidth]{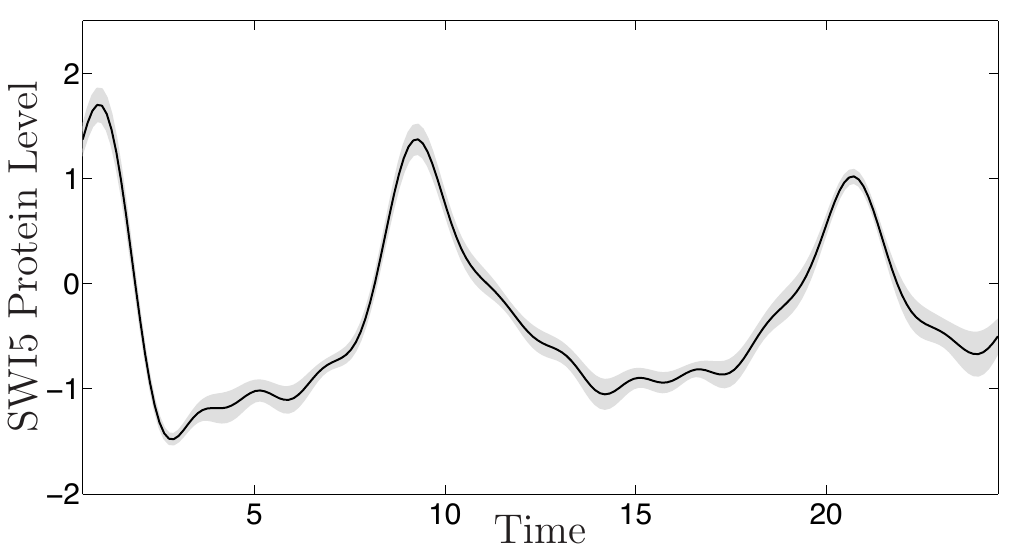}}\label{fig:swi5:factor:protein}}
\end{center}
\caption{Gene expresion profile and inferred protein concentration for
  ACE2 and SWI5. The first column shows the gene expression level. The
  second column shows the mean posterior over the transcription factor
  $u_q(t)$ and two standard deviations for the uncertainty.}
\label{fig:yeastSpellman}
\end{figure}

As an example, we tested the PITC approximation for the multiple
output Gaussian process using the benchmark yeast cell cycle dataset
of \citet{Spellman:yeastcellcy98}. Data is preprocessed as described in
\citet{Sanguinetti:probabilistic:tf:2006} with a final dataset of
$D=1975$ genes and $Q=104$ transcription factors. The data also
contains information about the structure of the network, basically a
matrix of connectivities between transcription factors and genes. This
is a matrix of size 1975$\times$104, where each entry is either a 0 or
a 1, indicating the absence or presence of a link between the gene and
the transcription factor protein. There are $N=24$ time points for
each gene. For the PITC approximation, we used $K=15$ fixed inducing
points, equally spaced in the input range. We optimize the
approximated marginal likelihood through scaled conjugate gradient
using $1000$ iterations, where each iteration takes about $0.72$
minutes.  Figure \ref{fig:ace2:gene:expression} shows the expresion
level $f_d(t)$ for ACE2 and in figure \ref{fig:ace2:factor:protein}
the inferred transcription factor $u_q(t)$. Equally, figure
\ref{fig:swi5:gene:expression} shows the expresion level $f_d(t)$ for
SWI5 and in figure \ref{fig:swi5:factor:protein} the inferred
transcription factor $u_q(t)$. The resulting shape of the transcription
factors can be seen as offset versions of the shape of the gene
expression data, which is a feature in this kind of networks.

\begin{figure}[ht!]
\begin{center}
  \subfigure[SNR for gene sensitivities associated to ACE2.]{
    {\includegraphics[width=0.48\textwidth]{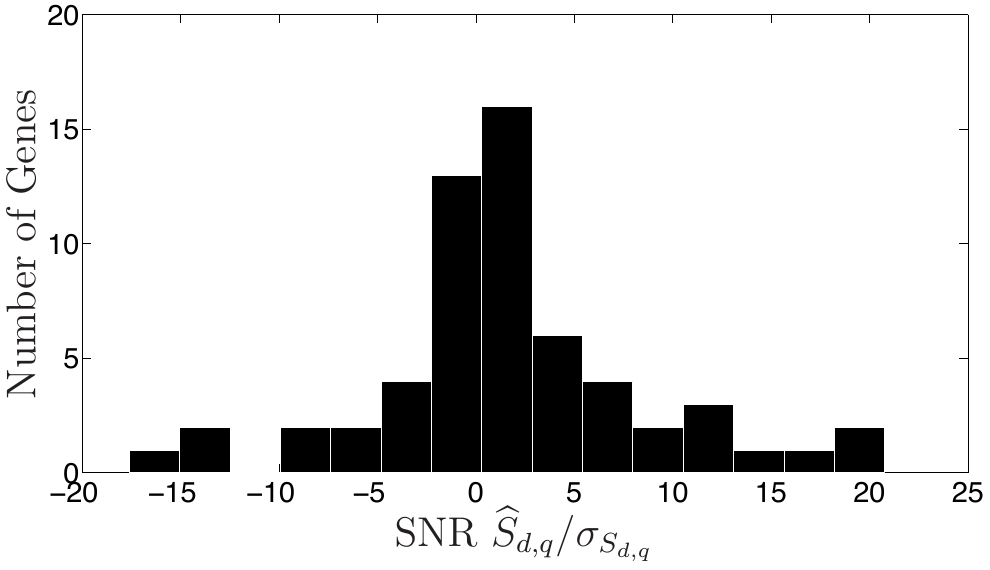}}\label{fig:yeastSpellmanHistPITC2}}
  \subfigure[SNR for gene sensitivities associated to SWI5.]{
    {\includegraphics[width=0.48\textwidth]{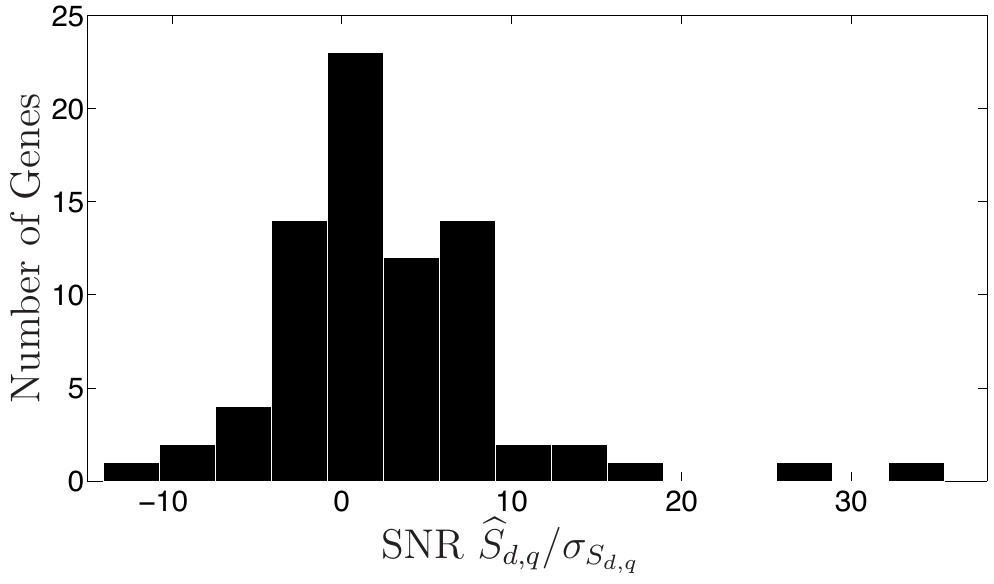}}\label{fig:yeastSpellmanHistPITC93}}
\end{center}
\caption{Histograms for the gene specific activities associated to
  ACE2 and SWI5. In (a), the SNR of gene-sensitivities associated to
  transcription factor ACE2 and in (b), the SNR of gene-sensitivities
  associated to SWI5.}
\label{fig:yeastSpellman}
\end{figure}

We can also use the sensitivity parameters $S_{d,q}$ for ranking the
relative influence of a particular transcription factor $q$, over a
particular gene $d$. In more detail, we use the signal-to-noise ratio
(SNR) $\widehat{S}_{d,q}/\sigma_{S_{d,q}}$, defined as the point
estimate $\widehat{S}_{d,q}$ of the sensitivity parameter $S_{d,q}$,
obtained after the optimization procedure, over the standard deviation
for that sentitiviy $\sigma_{S_{d,q}}$. An ad-hoc method to estimate
this standard deviation consists of approximating the mode of the
posterior density for the parameters $S_{d,q}$ with a second order
Taylor expansion: this is known as Laplace's approximation. For
details, the reader is referred to appendix \ref{laplace:approx}.
Figure \ref{fig:yeastSpellmanHistPITC2} shows a histogram of the
values of the signal-to-noise ratio of the sensitivities of all genes
in the dataset with respect to ACE2, this is, all the genes that,
according to the connectivity matrix, are regulated by ACE2. Some of
the highest SNR values are obtained for genes CTS1, SCW11, DSE1 and
DSE2, while, for example, NCE4 appears to be repressed with a low SNR
value. Similar results have been reported in other studies
\citep{Spellman:yeastcellcy98,Sanguinetti:probabilistic:tf:2006}.

Figure \ref{fig:yeastSpellmanHistPITC93} shows the signal-to-noise
ratio for the sensitivities of all the genes associated to the
transcription factor SWI5. Among others, genes AMN1 and PLC2 appear to
be activated by SWI5, as it has been confirmed experimentally by
\citet{ColmanLerner:swi5:2001}.

\section{Conclusions}\label{section:conclusions}

We have presented several sparse approximations for multiple output
GPs, in the context of convolution processes. Using these approximations
we can capture the correlated information among
outputs while reducing the amount of computational load for prediction
and optimization purposes. The computational complexity for the DTC and the FITC approximations is $O(NDK^2)$. 
The reduction in computational complexity for the PITC approximation is from $O(N^3D^3)$ to $O(N^3D)$. This
matches the computational complexity for modeling with independent GPs. However, as we have seen, 
the predictive power of independent GPs is lower. Also, since PITC makes a better approximation of the likelihood, 
the variance of the results is usually lower and approaches closely to the performance of the full GP, when compared to
DTC and FITC.     

With an appropriate selection of the kernel smoothing function we have
an indirect way to generate different forms for the covariance
function in the multiple output setup. We showed examples with
Gaussian kernels, for which a suitable standardization of the kernels
can be made, leading to competitive results in high-dimensional input
regression problems, as seen in the school exam score prediction
problem. The authors are not aware of other work in which this
convolution process framework has been applied in problems with high
input dimensions. Likewise, convolution appears naturally when solving
differential equations in dynamical systems and we showed how the
sparse methods can be applied to a large scale network inference
problem. In general, we do not have access to the connectivity matrix,
so to use this model in those situations we need to put sparse
priors over the sensitivity parameters. However, our motivation was to
show an example where sparse methods like the ones we proposed are
needed. We obtained sensible results that agree with previous
literature.

Recently, \citet{Titsias:variational09} highlighted how approximations
like FITC or PITC can exhibit a tendency to overfit when inducing
inputs are optimized. \citet{Titsias:variational09} proposed a
variational method with an associated lower bound to overcome to some
extent the overfitting problem. Following the ideas presented here, we
can combine easily the method of \citet{Titsias:variational09} and
propose a lower bound for the multiple output case.  This is part of
the future work.

\subsection*{Acknowledgements}

The authors would like to thank Edwin Bonilla for his valuable
feedback with respect to the exam score prediction example. The work
has benefited greatly from discussions with David Luengo, Michalis
Titsias, and Magnus Rattray. The authors are very grateful for support
from a Google Research Award ``Mechanistically Inspired Convolution
Processes for Learning'' and the EPSRC Grant No EP/F005687/1 ``Gaussian
Processes for Systems Identification with Applications in Systems
Biology''.


\appendix

\section{ Derivatives for the sparse methods}\label{app:theorem}
In this appendix, we present the derivatives needed to apply the
gradient methods in the optimization routines.  We present the first
order derivatives of the log-likelihood with respect to
$\boldK_{\boldf,\boldf}$, $\boldK_{\boldu,\boldf}$ and
$\boldK_{\boldu,\boldu}$. These derivatives can be combined with the
derivatives of $\boldK_{\boldf,\boldf}$, $\boldK_{\boldu,\boldf}$ and
$\boldK_{\boldu,\boldu}$ with respect to $\params$ and employ these
expressions in a gradient-like optimization procedure.

We also present the expressions for the Hessian matrix in Laplace's
approximation employed to compute the uncertainty of the sensitivity
parameters in the yeast cell cycle example.

We follow the notation of \citet{Brookes:matrix05} obtaining similar
results to \citet{Lawrence:larger07}. This notation allows us to apply the
chain rule for matrix derivation in a straight-forward manner. Let's
define $\mathbf{G}\veC=\vecO\mathbf{G}$, where $\vecO$ is the
vectorization operator over the matrix $\mathbf{G}$. For a function
$\mathcal{L}$ the equivalence between
$\frac{\partial\mathcal{L}}{\partial\mathbf{G}}$ and
$\frac{\partial\mathcal{L}}{\partial\mathbf{G}\veC}$ is given
through
$\frac{\partial\mathcal{L}}{\partial\mathbf{G}\veC}=
\left(\left(\frac{\partial\mathcal{L}}{\partial\mathbf{G}}\right)\veC\right)^\top$.
\subsection{First Derivatives of the log-likelihood for the gradient methods}\label{subsection:derivatives:PITC:FITC}
The obtain the hyperparameters, we maximize the following log-likelihood function,
\begin{align}
\mathcal{L}(\mathbf{Z},\bm{\theta})\propto
-\frac{1}{2}\log\abs{\mathbf{D}+\mathbf{K_{f,u}}\mathbf{K}^{-1}_{\mathbf{u,u}}\mathbf{K_{u,f}}}
-\frac{1}{2}\tr\left[
\left(\mathbf{D}+\mathbf{K_{f,u}}\mathbf{K}^{-1}_{\mathbf{u,u}}\mathbf{K_{u,f}}\right)^{-1}\mathbf{y}\mathbf{y}^\top\right]
\label{eq:log:likelihood:sparse}
\end{align}
where we have redefined $\mathbf{D}$ as
$\mathbf{D}=\left[\mathbf{K_{f,f}}-\mathbf{K_{f,u}}\mathbf{K}^{-1}_{\mathbf{u,u}}\mathbf{K_{u,f}}\right]\odot
\mathbf{M}+\bm{\Sigma}$,
to keep a simpler notation. Using the matrix inversion lemma and its
equivalent form for determinants, expression
\eqref{eq:log:likelihood:sparse} can be written as
\begin{align}
\mathcal{L}(\mathbf{Z},\bm{\theta})\propto& \frac{1}{2}\log
\abs{\mathbf{K_{u,u}}}-\frac{1}{2}\log\abs{\mathbf{A}}-\frac{1}{2}\log\abs{\mathbf{D}}
-\frac{1}{2}\tr\left[
\mathbf{D}^{-1}\mathbf{y}\mathbf{y}^\top\right]\nonumber\\
&+\frac{1}{2}\tr\left[\mathbf{D}^{-1}\mathbf{K_{f,u}}\mathbf{A}^{-1}\mathbf{K_{u,f}}\mathbf{D}
^{-1}\mathbf{y}\mathbf{y}^\top\right]\nonumber.
\end{align}
We can find $\frac{\partial \mathcal{L}}{\partial\bm{\theta}}$ and
$\frac{\partial \mathcal{L}}{\partial\mathbf{Z}}$ applying the chain
rule to $\mathcal{L}$ obtaining expressions for $\frac{\partial
\mathcal{L}}{\partial\mathbf{K_{f,f}}}$, $\frac{\partial
\mathcal{L}}{\partial\mathbf{K_{f,u}}}$ and $\frac{\partial
\mathcal{L}}{\partial\mathbf{K_{u,u}}}$ and combining those with the
relevant derivatives of the covariances wrt $\bm{\theta}$ and
$\mathbf{Z}$,
\begin{align}
\frac{\partial\mathcal{L}}{\partial\mathbf{G}\veC} =
\frac{\partial\mathcal{L}_\mathbf{A}}{\partial\mathbf{A}\veC}
\frac{\partial\mathbf{A}\veC}{\partial\mathbf{D}\veC}\frac{\partial\mathbf{D}\veC}{\partial\mathbf{G}\veC}+
\frac{\partial\mathcal{L}_\mathbf{D}}{\partial\mathbf{D}\veC}
\frac{\partial\mathbf{D}\veC}{\partial\mathbf{G}\veC}+
\left[\frac{\partial\mathcal{L}_\mathbf{A}}{\partial\mathbf{A}\veC}\frac{\partial\mathbf{A}\veC}{\partial\mathbf{G}\veC}
+\frac{\partial\mathcal{L}_\mathbf{G}}{\partial\mathbf{G}\veC}\right]\delta_{GK}\label{eq:chain:rule},
\end{align}
where the subindex in $\mathcal{L}_\mathbf{E}$ stands for those
terms of $\mathcal{L}$ which depend on $\mathbf{E}$, $\mathbf{G}$ is
either $\mathbf{K_{f,f}}$, $\mathbf{K_{u,f}}$ or $\mathbf{K_{u,u}}$
and $\delta_{GK}$ is zero if $\mathbf{G}$ is equal to
$\mathbf{K_{f,f}}$ and one in other case. Next we present
expressions for each partial derivative
\begin{gather*}
\begin{split}
\frac{\partial\mathcal{L}_\mathbf{A}}{\partial\mathbf{A}\veC}=
-\frac{1}{2}\left(\mathbf{C}\veC\right)^\top, \quad
\frac{\partial\mathbf{A}\veC}{\partial\mathbf{D}\veC}&=-\left(\mathbf{K_{u,f}}
\mathbf{D}^{-1}\otimes\mathbf{K_{u,f}}\mathbf{D}^{-1}\right),\quad
\frac{\partial\mathcal{L}_\mathbf{D}}{\partial\mathbf{D}\veC}=
-\frac{1}{2}\left(\left(\mathbf{D}^{-1}\mathbf{H}\mathbf{D}^{-1}\right)\veC\right)^\top\label{eq:grads:def:CH}
\end{split}
\\
\begin{split}
\frac{\partial\mathbf{D}\veC}{\partial\mathbf{K_{f,f}}\veC}=\diag(\mathbf{M}\veC),\quad
\frac{\partial\mathbf{D}\veC}{\partial\mathbf{K_{u,f}}\veC}=-\diag(\mathbf{M}\veC)\left[\left(\mathbf{I}
\otimes\mathbf{K_{f,u}}\mathbf{K}^{-1}_\mathbf{u,u}\right)+
\left(\mathbf{K_{f,u}}\mathbf{K}^{-1}_\mathbf{u,u}\otimes\mathbf{I}\right)\mathbf{T_D}\right],\label{eq:deriv:TD}
\end{split}
\\
\begin{split}
\frac{\partial\mathbf{D}\veC}{\partial\mathbf{K_{u,u}}\veC}=\diag(\mathbf{M}\veC)\left(\mathbf{K_{f,u}}\mathbf{K}^{-1}
_\mathbf{u,u}\otimes\mathbf{K_{f,u}}\mathbf{K}^{-1}_\mathbf{u,u}\right),
\frac{\partial\mathbf{A}\veC}{\partial\mathbf{K_{u,f}}\veC}=
\left(\mathbf{K_{u,f}}\mathbf{D}^{-1}\otimes\mathbf{I}\right)+
\left(\mathbf{I}\otimes\mathbf{K_{u,f}}\mathbf{D}^{-1}\right)\mathbf{T_A}
\label{eq:deriv:TA}
\end{split}
\\
\begin{split}
\frac{\partial\mathbf{A}\veC}{\partial\mathbf{K_{u,u}}\veC}=&\;\mathbf{I},\quad
\frac{\partial\mathcal{L}_\mathbf{K_{u,f}}}{\partial\mathbf{K_{u,f}}\veC}=\left(\left(\mathbf{A}^{-1}
\mathbf{K_{u,f}}\mathbf{D}^{-1}\mathbf{y}\mathbf{y}^\top\mathbf{D}^{-1}\right)\veC\right)^\top,
\quad
\frac{\partial\mathcal{L}_\mathbf{K_{u,u}}}{\partial\mathbf{K_{u,u}}\veC}=
\frac{1}{2}\left(\left(\mathbf{K}^{-1}_\mathbf{u,u}\right)\veC\right)^\top,
\end{split}
\end{gather*}
where
$\mathbf{C}=\mathbf{A}^{-1}+\mathbf{A}^{-1}\mathbf{K_{u,f}}\mathbf{D}^{-1}
\mathbf{y}\mathbf{y}^\top\mathbf{D}^{-1}\mathbf{K_{f,u}}\mathbf{A}^{-1}$,
$\mathbf{H}=\mathbf{D}-\mathbf{y}\mathbf{y}^\top+
\mathbf{K_{f,u}}\mathbf{A}^{-1}\mathbf{K_{u,f}}\mathbf{D}^{-1}
\mathbf{y}\mathbf{y}^\top +
\left(\mathbf{K_{f,u}}\mathbf{A}^{-1}\mathbf{K_{u,f}}\mathbf{D}^{-1}
  \mathbf{y}\mathbf{y}^\top\right)^\top$ and $\mathbf{T_D}$ and
$\mathbf{T_A}$ are \emph{vectorized transpose matrices} \citep[see,
e.g.,][]{Brookes:matrix05} and we have not included their dimensions
to keep the notation clearer. We can replace the above expressions in
\eqref{eq:chain:rule} to find the corresponding derivatives, so
\begin{subequations}
\begin{align}
\frac{\partial\mathcal{L}}{\partial\mathbf{K_{f,f}}\veC} = &
\;\frac{1}{2}\left[\left(\left(\mathbf{C}\right)\veC\right)^\top\left(\mathbf{K_{u,f}}
\mathbf{D}^{-1}\otimes\mathbf{K_{u,f}}\mathbf{D}^{-1}\right)-\frac{1}{2}\left(\left(\mathbf{D}^{-1}\mathbf{H}\mathbf{D}^{-1}
\right)\veC\right)^\top\right]\diag(\mathbf{M}\veC)
\label{eq:dL:dKff:1}\\
=&-\frac{1}{2}\left(\left(\mathbf{D}^{-1}\mathbf{J}\mathbf{D}^{-1}\right)\veC\right)^\top\diag(\mathbf{M}\veC)=
-\frac{1}{2}\left(\diag(\mathbf{M}\veC)\left(\mathbf{D}^{-1}\mathbf{J}\mathbf{D}^{-1}\right)\veC\right)^\top
\label{eq:dL:dKff:2}\\
=&-\frac{1}{2}\left(\left(\mathbf{D}^{-1}\mathbf{J}\mathbf{D}^{-1}\odot\mathbf{M}\right)\veC\right)^\top=
-\frac{1}{2}\left(\mathbf{Q}\veC\right)^\top \label{eq:dL:dKff:3}
\end{align}
\end{subequations}
or simply
\begin{align*}
\frac{\partial\mathcal{L}}{\partial\mathbf{K_{f,f}}} = & -\frac{1}{2}\mathbf{Q},
\end{align*}
where $\mathbf{J}=\mathbf{H}-\mathbf{K_{f,u}CK_{u,f}}$ and
$\mathbf{Q}=\left(\mathbf{D}^{-1}\mathbf{J}\mathbf{D}^{-1}\odot\mathbf{M}\right)$.
We have used the property
$\left(\mathbf{B}\veC\right)^\top\left(\mathbf{F}\otimes\mathbf{P}\right)=\left(\left(\mathbf{P}^
    \top\mathbf{B}\mathbf{F}\right)\veC\right)^\top$ in
\eqref{eq:dL:dKff:1} and the property
$\diag(\mathbf{B}\veC)\mathbf{F}\veC=(\mathbf{B}\odot\mathbf{F})\veC$,
to go from \eqref{eq:dL:dKff:2} to \eqref{eq:dL:dKff:3}. We also have
\begin{align}
\begin{split}
\frac{\partial\mathcal{L}}{\partial\mathbf{K_{u,f}}\veC} = &
\;\frac{1}{2}\left(\mathbf{Q}\veC\right)^\top\left[\left(\mathbf{I}
\otimes\mathbf{K_{f,u}}\mathbf{K}^{-1}_\mathbf{u,u}\right)+
\left(\mathbf{K_{f,u}}\mathbf{K}^{-1}_\mathbf{u,u}\otimes\mathbf{I}\right)\mathbf{T_D}\right]-
\frac{1}{2}\left(\mathbf{C}\veC\right)^\top\\
&\left[\left(\mathbf{K_{u,f}}\mathbf{D}^{-1}\otimes\mathbf{I}\right)
+\left(\mathbf{I}\otimes\mathbf{K_{u,f}}\mathbf{D}^{-1}\right)\mathbf{T_A}\right]+
\left(\left(\mathbf{A}^{-1}
\mathbf{K_{u,f}}\mathbf{D}^{-1}\mathbf{y}\mathbf{y}^\top\mathbf{D}^{-1}\right)\veC\right)^\top
\end{split}
\label{eq:dL:dKuf:1}\\
=&\left(\left(\mathbf{K}^{-1}_\mathbf{u,u}\mathbf{K_{u,f}}
\mathbf{Q}-\mathbf{C}\mathbf{K_{u,f}}\mathbf{D}^{-1}
+\mathbf{A}^{-1}\mathbf{K_{u,f}}\mathbf{D}^{-1}\mathbf{y}\mathbf{y}^\top\mathbf{D}^{-1}\right)\veC\right)^\top
\nonumber\label{eq:dL:dKuf:2}
\end{align}
or simply
\begin{align*}
\frac{\partial\mathcal{L}}{\partial\mathbf{K_{u,f}}}&= \mathbf{K}^{-1}_\mathbf{u,u}\mathbf{K_{u,f}}
\mathbf{Q}-\mathbf{C}\mathbf{K_{u,f}}\mathbf{D}^{-1}
+\mathbf{A}^{-1}\mathbf{K_{u,f}}\mathbf{D}^{-1}\mathbf{y}\mathbf{y}^\top\mathbf{D}^{-1},
\end{align*}
where in \eqref{eq:dL:dKuf:1},
$\left(\mathbf{Q}\veC\right)^\top\left(\mathbf{F}\otimes\mathbf{I}\right)\mathbf{T_D}=\left(\mathbf{Q}\veC\right)^
\top\mathbf{T_D}\left(\mathbf{I}\otimes\mathbf{F}\right)=\left(\mathbf{T}^\top_\mathbf{D}\mathbf{Q}\veC\right)^
\top\left(\mathbf{I}\otimes\mathbf{F}\right)=\left(\mathbf{Q}\veC\right)^\top\left(\mathbf{I}\otimes\mathbf{F}\right)$.
A similar analysis is formulated for the term involving
$\mathbf{T_A}$. Finally, results for
$\frac{\partial\mathcal{L}}{\partial\mathbf{K_{u,f}}}$ and
$\frac{\partial\mathcal{L}}{\partial\bm{\Sigma}}$ are obtained
as
\begin{align}
\frac{\partial\mathcal{L}}{\partial\mathbf{K_{u,u}}}
=-\frac{1}{2}\left(\mathbf{K}^{-1}_\mathbf{u,u}-\mathbf{C}-\mathbf{K}^{-1}_\mathbf{u,u}\mathbf{K_{u,f}}
\mathbf{Q}\mathbf{K_{f,u}}\mathbf{K}^{-1}_\mathbf{u,u}\right),
\quad \frac{\partial\mathcal{L}}{\partial\bm{\Sigma}} =
-\frac{1}{2}\mathbf{Q}.\nonumber
\end{align}

\subsection{Laplace approximation for the sensitivities}\label{laplace:approx}

As an ad-hoc procedure to compute the uncertainty in the sensitivy parameters, we employ a Laplace approximation
\citep[see e.g. Chapter 4][]{Bishop:PRLM06}. In 
particular, consider $Q=1$ and denote by $\mathbf{s}\in\Re^{D}$ the vector of sensitivities with entries given by $S_d$.
The Laplace aproximation $q(\mathbf{s})$ for the random vector $\mathbf{s}$ follows
\begin{align*}
q(\mathbf{s})& = \mathcal{N}(\mathbf{s}|\mathbf{s}_0, \bm{\Gamma}_{\mathbf{s}_0}), 
\end{align*} 
where $\mathbf{s}_0$ corresponds to a mode of the log-marginal likelihood for the sparse approximation and 
$\bm{\Gamma}^{-1}_{\mathbf{s}_0}=-\nabla\nabla \mathcal{L}(\mathbf{Z},\mathbf{s}, \bm{\eta})|_{\mathbf{s}=\mathbf{s}_0}$ with 
$\bm{\eta}$ representing the set of parameters belonging to the vector $\params$ without including $\mathbf{s}$. 
Asumming that after the optimization procedure we find a proper value for $\mathbf{s}_0$, we need to compute 
$\bm{\Gamma}^{-1}_{\mathbf{s}_0}$. 

For simplicity, let us denote by $\mathbf{Q}_{\boldf,\boldf}$ the approximated covariance in the marginal likelihood, 
this is, $\mathbf{Q}_{\boldf,\boldf}=\mathbf{D}+\mathbf{K_{f,u}}\mathbf{K}^{-1}_{\mathbf{u,u}}\mathbf{K_{u,f}}$. 
The log-marginal likelihood is the given as      
\begin{align*}
\mathcal{L}(\mathbf{Z},\bm{\theta})\propto
-\frac{1}{2}\log\abs{\mathbf{Q}_{\boldf,\boldf}}
-\frac{1}{2}\left[\mathbf{y}^\top\mathbf{Q}^{-1}_{\boldf,\boldf}\mathbf{y}\right].
\end{align*}
The derivative $\frac{\partial \mathcal{L}(\mathbf{Z},\bm{\theta})}{\partial S_d}$ is equal to
\begin{align*}
\frac{\partial \mathcal{L}(\mathbf{Z},\bm{\theta})}{\partial S_d}=\frac{\partial\mathcal{L}}{\partial 
\mathbf{Q}_{\boldf,\boldf}\veC}\frac{\partial \mathbf{Q}_{\boldf,\boldf}\veC}{\partial S_d}, 
\end{align*}
where 
\begin{align*}
\frac{\partial\mathcal{L}}{\partial \mathbf{Q}_{\boldf,\boldf}\veC} = -\frac{1}{2}\left[\Big(\mathbf{Q}^{-1}_{\boldf, \boldf}
\Big)\veC - \Big(\mathbf{Q}^{-1}_{\boldf, \boldf}\mathbf{y}\mathbf{y}^{\top}\mathbf{Q}^{-1}_{\boldf, \boldf}\Big)\veC \right]^\top,
\end{align*}
where the inverse matrix $\mathbf{Q}^{-1}_{\boldf, \boldf}$ is computed using $\mathbf{Q}^{-1}_{\boldf, \boldf}=\mathbf{D}^{-1}-
\mathbf{D}^{-1}\boldK_{\boldf,\boldu}\mathbf{A}^{-1}\boldK_{\boldu,\boldf}\mathbf{D}^{-1}$.
We assume the sensitivities are independent random variables, so we only need to compute the elements in the diagonal 
of $\bm{\Gamma}_{\mathbf{s}_0}$. Thus  
\begin{align*}
\frac{\partial^2 \mathcal{L}(\mathbf{Z},\bm{\theta})}{\partial S^2_d}&=\frac{\partial}{\partial S_d}
\bigg[\frac{\partial\mathcal{L}}{\partial \mathbf{Q}_{\boldf,\boldf}\veC}\frac{\partial \mathbf{Q}_{\boldf,\boldf}\veC}
{\partial S_d} \bigg] = \frac{\partial\mathcal{L}}{\partial \mathbf{Q}_{\boldf,\boldf}\veC}
\frac{\partial^2 \mathbf{Q}_{\boldf,\boldf}\veC}{\partial S^2_d}+ \frac{\partial}{\partial S_d}
\bigg[\frac{\partial\mathcal{L}}{\partial \mathbf{Q}_{\boldf,\boldf}\veC}\bigg]\frac{\partial \mathbf{Q}_{\boldf,\boldf}\veC}
{\partial S_d}. 
\end{align*}
Finally, in the above expression, we need to compute 
\begin{align*}
\frac{\partial}{\partial S_d}
\bigg[\frac{\partial\mathcal{L}}{\partial \mathbf{Q}_{\boldf,\boldf}\veC}\bigg]&=\frac{\partial}{\partial S_d}\bigg\{
-\frac{1}{2}\left[\Big(\mathbf{Q}^{-1}_{\boldf, \boldf}
\Big)\veC - \Big(\mathbf{Q}^{-1}_{\boldf, \boldf}\mathbf{y}\mathbf{y}^{\top}\mathbf{Q}^{-1}_{\boldf, \boldf}\Big)\veC \right]^\top
\bigg\}\\
&=-\frac{1}{2}\left[\frac{\partial}{\partial S_d}\Big(\mathbf{Q}^{-1}_{\boldf, \boldf}
\Big)\veC - \frac{\partial}{\partial S_d}\Big(\mathbf{Q}^{-1}_{\boldf, \boldf}\mathbf{y}\mathbf{y}^{\top}\mathbf{Q}^{-1}
_{\boldf, \boldf}\Big)\veC \right]^\top\\
&=-\frac{1}{2}\bigg\{\left[\frac{\dif}{\dif \mathbf{Q}_{\boldf,\boldf}\veC}\Big(\mathbf{Q}^{-1}_{\boldf, \boldf}
\Big)\veC - \frac{\dif}{\dif \mathbf{Q}_{\boldf,\boldf}\veC}\Big(\mathbf{Q}^{-1}_{\boldf, \boldf}\mathbf{y}
\mathbf{y}^{\top}\mathbf{Q}^{-1}
_{\boldf, \boldf}\Big)\veC \right]\frac{\partial \mathbf{Q}_{\boldf,\boldf}\veC}{\partial S_d}\bigg\}^\top\\
&=\frac{1}{2}\bigg\{\left[\mathbf{Q}^{-1}_{\boldf, \boldf}\otimes \mathbf{Q}^{-1}_{\boldf, \boldf}
- \mathbf{Q}^{-1}_{\boldf, \boldf}\mathbf{y}\mathbf{y}^{\top}\mathbf{Q}^{-1}_{\boldf, \boldf}\otimes\mathbf{Q}^{-1}_{\boldf, \boldf}
- \mathbf{Q}^{-1}_{\boldf, \boldf}\otimes\mathbf{Q}^{-1}_{\boldf, \boldf}\mathbf{y}\mathbf{y}^{\top}\mathbf{Q}^{-1}_{\boldf, \boldf}
\right]\frac{\partial \mathbf{Q}_{\boldf,\boldf}\veC}{\partial S_d}\bigg\}^\top.
\end{align*}
We do not need to compute the Kronecker products above. Instead, we use the property $(\mathbf{PBF})\veC = 
(\mathbf{F}^\top\otimes\mathbf{P})\mathbf{B}\veC$, leading to 
\begin{align*}
\frac{\partial}{\partial S_d}
\bigg[\frac{\partial\mathcal{L}}{\partial \mathbf{Q}_{\boldf,\boldf}\veC}\bigg]&=
\frac{1}{2}\bigg[\left(\mathbf{Q}^{-1}_{\boldf, \boldf}\frac{\partial \mathbf{Q}_{\boldf,\boldf}}{\partial S_d}
\mathbf{Q}^{-1}_{\boldf, \boldf}
- \mathbf{Q}^{-1}_{\boldf, \boldf}\mathbf{y}\mathbf{y}^{\top}\mathbf{Q}^{-1}_{\boldf, \boldf}
\frac{\partial \mathbf{Q}_{\boldf,\boldf}}{\partial S_d}\mathbf{Q}^{-1}_{\boldf, \boldf}
- \mathbf{Q}^{-1}_{\boldf, \boldf}\frac{\partial \mathbf{Q}_{\boldf,\boldf}}{\partial S_d}
\mathbf{Q}^{-1}_{\boldf, \boldf}\mathbf{y}\mathbf{y}^{\top}\mathbf{Q}^{-1}_{\boldf, \boldf}
\right)\veC\bigg]^\top.
\end{align*}

\bibliography{reportbiblio}

\end{document}